\documentclass[journal]{IEEEtran}
\usepackage[T1]{fontenc}
\usepackage[utf8]{inputenc}
\usepackage{textcomp}
\usepackage{tgtermes}

\usepackage{amsmath,amsfonts,amsthm,amssymb,mathtools,cases}

\usepackage{newtxmath}
\usepackage{bm}
\usepackage{mathrsfs}
\usepackage{upgreek}

\newcommand{\symbffrak}[1]{\bm{\mathfrak{#1}}} % Bold Fraktur
\newcommand{\symfrak}[1]{\mathfrak{#1}} % Bold Fraktur
\newcommand{\symcal}[1]{\mathcal{#1}} % Calligraphic
\newcommand{\symbfcal}[1]{\bm{\mathscr{#1}}} % Bold calligraphic
\newcommand{\symbf}[1]{\bm{\mathrm{#1}}} % Bold upright
\newcommand{\symbb}[1]{\mathbb{#1}} % Blackboard bold
\newcommand{\dprime}{^{\prime\prime}}
\usepackage{graphicx}
\graphicspath{{./img/}}
\DeclareGraphicsExtensions{.png,.pdf}
\usepackage{xcolor}
\usepackage{url}

\usepackage[%
style = ieee,%
citestyle = numeric-comp,%
bibencoding = utf8,%
datamodel = software,%
backend = biber,%
giveninits = true,%
abbreviate = true,
maxbibnames = 10,%
maxcitenames = 2,%
sortcites = true,%
doi = true,%
isbn = false,%
url = false,%
eprint = true,%
]{biblatex}

\usepackage{pgfplots}
\pgfplotsset{compat=newest}
\usepgflibrary{plotmarks}
\usetikzlibrary{
  tikzmark,
  calc,
  arrows,
  chains,
  decorations.pathreplacing,
  decorations.pathmorphing,
  shapes,
  matrix,
}

\newcommand{\ArrayLength}{10}
\newcommand{\ArrayHeight}{2}
\newcommand{\Scale}{1.5}
\newcommand{\PHeight}{0.5}
\newcommand{\SolveHeight}{1.9}

\usepackage{algorithm, algpseudocode}
\makeatletter
\renewcommand{\ALG@beginalgorithmic}{\footnotesize}

\algrenewcommand{\algorithmicrequire}{\textbf{Input:}}
\algrenewcommand{\algorithmicensure}{\textbf{Output:}}

\algrenewcommand{\algorithmiccomment}[1]{%
  \hfill{\footnotesize\ttfamily /* #1 */}%
}

\algnewcommand{\LineComment}[1]{%
  \Statex \hskip\ALG@thistlm {\footnotesize\ttfamily /* #1 */}%
}

\algnewcommand{\LineCommentAfterFunction}[1]{%
  \Statex \hskip\ALG@tlm {\footnotesize\ttfamily /* #1 */}%
}

\algnewcommand{\LongEq}[1]{%
  \State \parbox[t]{\dimexpr \linewidth - \ALG@thistlm}{#1}%
}

\makeatother

\definecolor{cadetblue}{rgb}{0.37, 0.62, 0.63}
\definecolor{burntorange}{rgb}{0.8, 0.33, 0.0}
\definecolor{americanrose}{rgb}{1.0, 0.01, 0.24}
\definecolor{applegreen}{rgb}{0.55, 0.71, 0.0}
\definecolor{darkmagenta}{rgb}{0.55, 0.0, 0.55}
\definecolor{peru}{rgb}{0.80, 0.52, 0.25}
\definecolor{navy}{rgb}{0.0, 0.0, 0.5}
\definecolor{maroon}{rgb}{0.5, 0.0, 0.0}
\definecolor{gold}{rgb}{1.0, 0.84, 0.0}
\definecolor{crimson}{rgb}{0.86, 0.08, 0.24}
\definecolor{darkgreen}{rgb}{0., 0.3, 0.}

\usepackage{newfloat}
\usepackage{subcaption}

\usepackage[inline]{enumitem}
\usepackage{etoolbox}
\usepackage{xspace}
\usepackage{siunitx}

\usepackage[mathlines, switch]{lineno}

\newcommand*\patchAmsMathEnvironmentForLineno[1]{%
  \expandafter\let\csname old#1\expandafter\endcsname\csname #1\endcsname
  \expandafter\let\csname oldend#1\expandafter\endcsname\csname end#1\endcsname
  \renewenvironment{#1}%
  {\linenomath\csname old#1\endcsname}%
  {\csname oldend#1\endcsname\endlinenomath}}%
\newcommand*\patchBothAmsMathEnvironmentsForLineno[1]{%
  \patchAmsMathEnvironmentForLineno{#1}%
  \patchAmsMathEnvironmentForLineno{#1*}}%
\AtBeginDocument{%
  \patchBothAmsMathEnvironmentsForLineno{equation}%
  \patchBothAmsMathEnvironmentsForLineno{align}%
  \patchBothAmsMathEnvironmentsForLineno{flalign}%
  \patchBothAmsMathEnvironmentsForLineno{alignat}%
  \patchBothAmsMathEnvironmentsForLineno{gather}%
  \patchBothAmsMathEnvironmentsForLineno{multline}%
}

\usepackage[hidelinks]{hyperref}

\newcommand{\IntegerP}{\symbb{N}}
\newcommand{\IntegerPP}{\symbb{N}_*}
\newcommand{\Real}{\symbb{R}}

\newcommand{\RealPP}{\symbb{R}_{++}}

\newcommand\given{{\mathbin{}\mid\mathbin{}}}
\newcommand\vect[1]{\symbf{#1}}

\providecommand\given{} % so it exists
\newcommand\SetSymbol[1][]{
  \nonscript\,#1\vert \allowbreak \nonscript\,\mathopen{}}
\DeclarePairedDelimiterX\Set[1]{\lbrace}{\rbrace}%
{ \renewcommand\given{\SetSymbol[\delimsize]} #1 }
\DeclarePairedDelimiterX\innerp[2]{\langle}{\rangle}{#1
  \mathop{}\delimsize\vert\mathop{} #2}
\DeclarePairedDelimiterX\norm[1]\lVert\rVert{\ifblank{#1}{\:\cdot\:}{#1}}

\DeclareMathOperator{\Span}{span}
\DeclareMathOperator{\trace}{trace}

\DeclareMathOperator{\diag}{diag}
\DeclareMathOperator{\Fix}{Fix}

\DeclareMathOperator{\expect}{\mathbb{E}}

\DeclareMathOperator{\kernel}{ker}

\DeclareMathOperator{\Id}{Id}

\DeclareMathOperator{\argmin}{argmin}

\usepackage{stackengine}
\stackMath

\makeatletter
\DeclareRobustCommand{\rvdots}{%
  \vbox{
    \baselineskip4\p@\lineskiplimit\z@
    \kern-\p@
    \hbox{.}\hbox{.}\hbox{.}
  }}
\makeatother

\usepackage{thmtools}
\usepackage[unq]{unique}

\declaretheoremstyle[%
  headfont=\normalfont\bfseries,
  notefont=\mdseries,
  notebraces={(}{)},
  bodyfont=\normalfont,
  postheadspace=1ex%,
]{mystyle}

\declaretheorem[style=mystyle,
  name=Theorem,
  refname={theorem,theorems},
  Refname={Theorem,Theorems},
]{thm}

\declaretheorem[style=mystyle,
  name=Lemma,
  refname={lemma,lemmata},
  Refname={Lemma,Lemmata},
  numberlike=thm,
]{lemma}

\declaretheorem[style=mystyle,
  name=Assumption,
  refname={assumption,assumptions},
  Refname={Assumption,Assumptions},
  numberlike=thm,
]{assumption}

\declaretheorem[style=mystyle,
  name=Assumptions,
  refname={assumptions},
  Refname={Assumptions},
  numberlike=thm,
]{assumptions}

\declaretheorem[style=mystyle,
  name=Contributions,
  refname={contributions},
  Refname={Contributions},
  numbered=no,
]{contributions}

\declaretheorem[style=mystyle,
  name=Algorithm,
  refname={algorithm,algorithms},
  Refname={algorithm,algorithms}
]{algo}

\declaretheorem[
  style=mystyle,
  name=Challenges,
  refname={challenges,challenges},
  Refname={Challenges,Challenges},
  numberlike=thm
]{challenges}

\newlist{thmlist}{enumerate}{1}
\setlist[thmlist]{label=\textbf{(\roman{*})}, ref=\thethm(\roman{*}), noitemsep}

\newlist{lemlist}{enumerate}{1}
\setlist[lemlist]{label=\textbf{(\roman{*})}, ref=\thelemma(\roman{*}), noitemsep}

\newlist{exlist}{enumerate}{1}
\setlist[exlist]{label=\textbf{(\roman{*})}, ref=\theexample(\roman{*}), noitemsep}

\newlist{factlist}{enumerate}{1}
\setlist[factlist]{label=\textbf{(\roman{*})}, ref=\thefact(\roman{*}), noitemsep}

\newlist{proplist}{enumerate}{1}
\setlist[proplist]{label=\textbf{(\roman{*})}, ref=\theprop(\roman{*}), noitemsep}

\newlist{asslist}{enumerate}{1}
\setlist[asslist]{label=\textbf{(\roman{*})}, ref=\theassumption(\roman{*}), noitemsep}

\newlist{assslist}{enumerate}{1}
\setlist[assslist]{label=\textbf{(\roman{*})}, ref=\theassumptions(\roman{*}), noitemsep}

\newlist{deflist}{enumerate}{1}
\setlist[deflist]{label=\textbf{(\roman{*})}, ref=\thedefinition(\roman{*}), noitemsep}

\newlist{algolist}{enumerate}{1}
\setlist[algolist]{label=\textbf{(\roman{*})}, ref=\thealgo(\roman{*}), noitemsep}

\newlist{claimlist}{enumerate}{1}
\setlist[claimlist]{label=\textbf{(\roman{*})}, ref=\theclaim(\roman{*}), noitemsep}

\newlist{applist}{enumerate}{1}
\setlist[applist]{label=\textbf{(\roman{*})},
  ref=\thesection(\roman{*}), noitemsep}

\newlist{contriblist}{enumerate}{1}
\setlist[contriblist]{label=\textbf{(\arabic{*})},
  ref=\thecontributions(\arabic{*}), noitemsep}

\newlist{MyEnumSec}{enumerate}{1}
\setlist[MyEnumSec]{label=\textbf{\thesection(\roman{*})},
  ref=Item~\thesection(\roman{*}), noitemsep}

\newlist{MyEnumSubSec}{enumerate}{1}
\setlist[MyEnumSubSec]{label=\textbf{\thesubsection(\roman{*})},
  ref=Item~\thesubsection(\roman{*}), noitemsep, wide = 0pt,
  leftmargin = *}

\newlist{challengelist}{enumerate}{1}
\setlist[challengelist]{%
  label=\textbf{(\roman*)},        % (i), (ii), ...
  ref=\thechallenges(\roman*),     % cross-reference format
  noitemsep
}

\usepackage[capitalize]{cleveref}

\crefname{thm}{Theorem}{Theorems}
\crefname{prop}{Proposition}{Propositions}
\crefname{assumption}{Assumption}{Assumptions}
\crefname{assumptions}{Assumptions}{Assumptions}
\crefname{contributions}{Contributions}{Contributions}
\crefname{challenges}{Challenges}{Challenges}
\crefname{lemma}{Lemma}{Lemmata}
\crefname{definition}{Definition}{Definitions}
\crefname{example}{Example}{Examples}
\crefname{algo}{Algorithm}{Algorithms}
\crefname{fact}{Fact}{Facts}
\crefname{claim}{Claim}{Claims}
\crefname{appendix}{Appendix}{Appendices}
\crefname{coroll}{Corollary}{Corollaries}
\crefname{figure}{Figure}{Figures}
\crefname{section}{Section}{Sections}

\crefname{thmlisti}{Theorem}{Theorems}
\crefname{lemlisti}{Lemma}{Lemmata}
\crefname{proplisti}{Proposition}{Propositions}
\crefname{asslisti}{Assumption}{Assumptions}
\crefname{assslisti}{Assumption}{Assumptions}
\crefname{contriblisti}{Contribution}{Contributions}
\crefname{deflisti}{Definition}{Definitions}
\crefname{exlisti}{Example}{Examples}
\crefname{algolisti}{Algorithm}{Algorithms}
\crefname{factlisti}{Fact}{Facts}
\crefname{claimlisti}{Claim}{Claims}
\crefname{applisti}{Appendix}{Appendices}
\crefname{challengelisti}{Challenge}{Challenges}
\crefname{MyEnumSeci}{}{}
\crefname{MyEnumSubSeci}{}{}

\makeatletter

\newcommand*{\ie}{%
  \@ifnextchar{,}%
  {\textit{i.e.}}%
  {\textit{i.e.,}\@\xspace}%
}
\newcommand*{\eg}{%
  \@ifnextchar{,}%
  {\textit{e.g.}}%
  {\textit{e.g.,}\@\xspace}%
}
\newcommand*{\etc}{%
  \@ifnextchar{.}%
  {\textit{etc}}%
  {\textit{etc.}\@\xspace}%
}
\newcommand*{\etal}{%
  \@ifnextchar{.}%
  {\textit{et al}}%
  {\textit{et al.}\@\xspace}%
}
\newcommand*{\cf}{%
  \@ifnextchar{.}%
  {\textit{cf}}%
  {\textit{cf.}\@\xspace}%
}
\newcommand*{\aka}{%
  \@ifnextchar{,}%
  {\textit{a.k.a.}}%
  {\textit{a.k.a.}\@\xspace}%
}
\newcommand*{\whp}{%
  \@ifnextchar{,}%
  {\textit{w.h.p.}}%
  {\textit{w.h.p.}\@\xspace}%
}
\makeatother

\makeatletter
\newlength{\negph@wd}
\DeclareRobustCommand{\negphantom}[1]{%
  \ifmmode
    \mathpalette\negph@math{#1}%
  \else
    \negph@do{#1}%
  \fi
}
\newcommand{\negph@math}[2]{\negph@do{$\m@th#1#2$}}
\newcommand{\negph@do}[1]{%
  \settowidth{\negph@wd}{#1}%
  \hspace*{-\negph@wd}%
}
\makeatother

\allowdisplaybreaks
\begin{document}

\title{Nonparametric Bellman Mappings for Value Iteration in Distributed Reinforcement Learning}

\author{Yuki~Akiyama and Konstantinos~Slavakis\IEEEauthorrefmark{1}%
  \thanks{\IEEEauthorrefmark{1}K.~Slavakis is with the Institute of Science Tokyo, Department of Information and
    Communications Engineering, 4259-G2-4 Nagatsuta-Cho, Midori-Ku, Yokohama, Kanagawa, 226-8502 Japan. Email:
    \texttt{slavakis@ict.eng.isct.ac.jp}.%% akiyamayuki1010@gmail.com
  }
}

\maketitle
% \linenumbers

\begin{abstract}
  This paper introduces novel Bellman mappings (B-Maps) for value iteration (VI) in distributed reinforcement learning
  (DRL), where agents are deployed over an undirected, connected graph/network with arbitrary topology---but without a
  centralized node, that is, a node capable of aggregating all data and performing computations. Each agent constructs a
  nonparametric B-Map from its private data, operating on Q-functions represented in a reproducing kernel Hilbert space,
  with flexibility in choosing the basis for their representation. Agents exchange their Q-function estimates only with
  direct neighbors, and unlike existing DRL approaches that restrict communication to Q-functions, the proposed
  framework also enables the transmission of basis information in the form of covariance matrices, thereby conveying
  additional structural details. Linear convergence rates are established for both Q-function and covariance-matrix
  estimates toward their consensus values, regardless of the network topology, with optimal learning rates determined by
  the ratio of the smallest positive eigenvalue (the graph's Fiedler value) to the largest eigenvalue of the graph
  Laplacian matrix. A detailed performance analysis further shows that the proposed DRL framework effectively
  approximates the performance of a centralized node, had such a node existed. Numerical tests on two benchmark control
  problems confirm the effectiveness of the proposed nonparametric B-Maps relative to prior methods. Notably, the tests
  reveal a counter-intuitive outcome: although the framework involves richer information exchange---specifically through
  transmitting covariance matrices as basis information---it achieves the desired performance at a lower cumulative
  communication cost than existing DRL schemes, underscoring the critical role of sharing basis information in
  accelerating the learning process.
\end{abstract}


\IEEEpeerreviewmaketitle

\section{Introduction}\label{sec:intro}

In reinforcement learning (RL), an agent interacts with and controls a system by making sequential decisions or actions
based on feedback from the surrounding environment~\cite{bertsekas1996neuro, bertsekas2019reinforcement,
  Sutton:IntroRL:18}. This feedback is typically provided in the form of an \textit{one-step loss}\/ $g(\cdot)$ or,
equivalently, a reward defined as $-g(\cdot)$~\cite{bertsekas2019reinforcement, Sutton:IntroRL:18}. The agent uses this
feedback to learn an \textit{optimal policy}\/ $\mu_*(\cdot)$, a function or strategy that prescribes actions based on
the system's state, to minimize the \textit{long-term loss/penalty}\/---also known as the Q-function. The Q-function
represents the total penalty the agent would incur if all future decisions were made according to $\mu_*(\cdot)$. The
Bellman mapping (B-Map) is a fundamental tool for computing Q-functions, with its fixed points playing a crucial role in
determining $\mu_*(\cdot)$~\cite{bertsekas2019reinforcement}. Over the years, various B-Map formulations and algorithmic
approaches have been developed to model and compute Q-functions. These include classical
Q-learning~\cite{Sutton:IntroRL:18, bertsekas2019reinforcement} as well as methods that employ functional
approximation~\cite{ormoneit2002kernel, lagoudakis2003lspi, Nedic:LSPE:03, bertsekas2004improved, xu2007klspi,
  Bae:kerneltd1:11, onlinebrloss:16, regularizedpi:16, akiyama2024nonparametric}.

This work contributes to distributed reinforcement learning (DRL) and, more broadly, distributed
learning~\cite{sayed2014adaptation}, where agents are deployed over a network or graph, each associated with a network
node~\cite{distributed.DDP, distributed.DDP.path, distributed.Gossip, distributed.ActorCritic, distributed.LSTD}. The
DRL premise offers notable advantages over non-distributed RL, especially in scenarios where a single agent cannot
process all available data---whether due to privacy considerations or computational limitations (\eg, data
centers)---thereby requiring the data as well as the workload to be distributed across multiple computing platforms,
\eg, \cite{Wang:DistributedRL:2022, Wang:2025:DistRL}. It is also well-suited to multi-task
RL---that is, situations in which each agent handles a variety of tasks that must be performed in parallel, with agents
communicating and coordinating efficiently to learn a global policy that ensures the successful completion of all
tasks~\cite{Hendawy:MOORE:ICLR24}.

Although a detailed description of the DRL setting is provided in \cref{ass:setting}, the basic premise can be
summarized as follows: a centralized node capable of gathering all data and performing the required computations on
behalf of the agents is not available. Instead, the agents interact with a shared environment without exchanging any
state-action information, independently compute their own Q-function estimates, and communicate these estimates only to
their immediate neighbors. Through such localized communication, the agents collectively converge to a network-wide
consensus Q-function, which in turn enables the identification of optimal policies across the network.

A plethora of studies have explored RL tasks across networks of agents, and the term multi-agent RL (MARL) is often used
as a blanket designation for this diverse body of work. DRL substantially overlaps with MARL, and the distinction
between the two is frequently blurred. The primary goal in MARL remains the computation of a network-wide Q-function
using information collected across the network, with many MARL approaches assuming that agents share their state
information~\cite{distributed.Q, distributed.LSTD.ADMM, distributed.TD, distributed.LSTD.primal, distributed.deep}. For
instance, in cooperative MARL, state-action information may be shared with all agents via a fusion
node~\cite{Hsu:NeurIPS24}, or one agent's action may directly influence another agent's state
transition~\cite{Du:AAAI24}. In contrast, the setting considered in this paper (see \cref{ass:setting}) ensures that
each agent's state-action information remains private. Some MARL methods, such as~\cite{distributed.LSTD.primal}, can
operate in environments where state-space information is either shared or private. It is also common to assume that each
agent $n$, with $n \in \{1, 2, \ldots, N \}$ and $N$ denoting the number of agents, has access only to its own one-step
loss function $g^{(n)}(\cdot)$, without knowledge of other agents' losses,
\eg,~\cite{distributed.LSTD.primal}. Furthermore, in federated RL, state-action information stays private, but agents
are deployed over a network with a star topology, where a central fusion node aggregates their transmitted information
and redistributes it back, \eg, \cite{Zhang:2024:FedSARSA, Lan:2025:AFedPG}. Unlike federated RL, the present setting
(\cref{ass:setting}) imposes no restrictions on the network topology, and all established performance results
(\cref{sec:performance}) hold regardless of the adopted topology. To maintain a focused, concise, and coherent
presentation of the proposed framework, extensions of \cref{ass:setting} and the associated algorithmic developments to
more general MARL settings, or a detailed treatment of the specific federated RL problem, are deferred to future
work. These extensions may consider scenarios in which agents share state-action information locally, while extensions
to multi-task RL are also reserved for future investigation. Given the conceptual overlap, the lack of a strict boundary
between DRL and MARL, and the broader connotation of the term ``distributed,'' particularly in connection with
distributed learning and optimization~\cite{sayed2014adaptation, distributed.DDP}, the term DRL is retained throughout
this work.

Classical Q-learning has been extended into DRL, where, as is often the case in Q-learning, the state space is
considered discrete rather than continuous, and Q-functions are represented in a tabular form~\cite{distributed.Q,
  distributed.DDP, distributed.DDP.path}. Similarly, standard B-Maps have also been adapted for DRL
in~\cite{distributed.DDP, distributed.DDP.path}, but the state space remains discrete, and the network or graph topology
depends on the specific state space. To overcome the limitations of tabular Q-functions and discrete state spaces,
functional approximation models for Q-functions have been explored, particularly in deep-learning-based
DRL~\cite{distributed.A3C, distributed.deep}, where a centralized node is assumed to exist within the network. DRL with
functional approximation has also been developed for fully distributed settings, eliminating the need for a centralized
node, or even partial state-information sharing among agents~\cite{distributed.ActorCritic, distributed.Gossip,
  distributed.LSTD, distributed.DDP, distributed.DDP.path, distributed.A3C,
  distributed.LSTD.primal}. Study~\cite{NN.gossip} can accommodate general functional approximation models for
Q-functions, including neural networks, and can be applied to address general DRL tasks.

This paper builds upon the recently introduced nonparametric B-Maps~\cite{akiyama2024nonparametric} and extends them to
DRL\@. Specifically, the B-Maps considered here operate on Q-functions within a reproducing kernel Hilbert space
(RKHS)~\cite{aronszajn1950, scholkopf2002learning}, leveraging both the functional approximation capabilities of RKHS,
such as the universal-approximation properties of Gaussian kernels~\cite{Gyorfi:DistrFree:10, chen96ebf}, and the
reproducing property of its inner product. This contrasts with standard B-Maps, in which Q-functions are treated as
elements of a Banach space lacking an inner-product structure~\cite{bertsekas2019reinforcement, ormoneit2002kernel}. By
selecting an RKHS as the functional approximation space for Q-functions, the proposed approach becomes fully
nonparametric~\cite{Gyorfi:DistrFree:10}, eliminating the need for statistical priors or assumptions on the data while
minimizing user-induced modeling bias. A key trade-off of this distribution-free approach is that the number of free
parameters required to represent Q-function estimates grows with the size of the dataset. To mitigate this, a
dimensionality-reduction strategy based on random Fourier features (RFFs)~\cite{rff} is employed, reducing the impact of
the ``curse of dimensionality.''

In summary, this paper offers the following novel contributions to DRL\@.

\begin{contributions}\mbox{}
  \begin{contriblist}

  \item\label{contrib:Bmaps} \textbf{(Novel B-Maps for DRL)} The benefits of the nonparametric B-Maps introduced
    in~\cite{akiyama2024nonparametric} are extended here to the DRL setting. Each agent $n$ has the flexibility to
    select its own basis functions $\vect{\Psi}^{(n)}$ within the ambient RKHS, thereby enabling the construction of
    \textit{agent-specific}\/ B-Maps.

  \item\label{contrib:share.cov.mat} \textbf{(Exchange of covariance-matrix data)} Unlike prior
    methods~\cite{distributed.ActorCritic, distributed.Gossip, distributed.LSTD, distributed.DDP, distributed.DDP.path,
      distributed.A3C, distributed.LSTD.primal, distributed.LSTD.ADMM, distributed.TD}, where agents communicate only
    their Q-function estimates, the proposed framework also enables the exchange of information about each agent's basis
    functions $\vect{\Psi}^{(n)}$ via covariance matrices.

  \item\label{contrib:performance.analysis} \textbf{(Performance analysis)} The framework guarantees
    linear convergence of both the nodal Q-function and covariance-matrix estimates toward their consensus values,
    irrespective of the network topology. The optimal learning rates for the iterative updates are determined by the
    ratio of the smallest positive eigenvalue (\aka the algebraic connectivity or Fiedler value of the
    graph~\cite{Bapat.graphs}) to the largest eigenvalue of the graph Laplacian matrix. Furthermore, a tunable bound is
    established on the deviation of each nodal Q-function estimate from the fixed point of a centralized nonparametric
    B-Map, had a centralized node existed. By adjusting a design parameter, this bound can be made arbitrarily small,
    indicating that the proposed DRL framework closely approximates the behavior of a centralized node.

  \end{contriblist}
\end{contributions}

Numerical tests reveal a surprising and counter-intuitive finding: although the proposed method, in principle, involves
exchanging \textit{more}\/ information among agents than previous approaches---specifically through the sharing of
covariance matrices (\cref{contrib:share.cov.mat})---it achieves the desired performance with a \textit{lower} overall
communication cost. This underscores the crucial role of basis information, whose exchange significantly accelerates the
learning process.

The remainder of the paper is organized as follows. \cref{sec:preliminaries} introduces notation and RL preliminaries,
while \cref{sec:setting} outlines the general assumptions of the DRL setting. A centralized nonparametric B-Map is
presented in \cref{sec:central.Bmap}, the challenges of distributed solutions in \cref{sec:challenges}, and the proposed
DRL approach in \cref{sec:distribute.Q,sec:distribute.cov,sec:algo}, followed by the performance analysis of the
proposed \cref{algo} in \cref{sec:performance}. Numerical tests are reported in \cref{sec:numerical}, conclusions in
\cref{sec:conclusions}, and detailed proofs supporting the analysis are collected in the appendix.

\section{Distributed Reinforcement Learning}\label{sec:DRL}

\subsection{Preliminaries and notation}\label{sec:preliminaries}

A number $N$ of agents are considered, distributed over a connected~\cite{Gross.GraphTheory.book} network/graph
$\symcal{G} \coloneqq (\symcal{N}, \symcal{E})$, with nodes $\symcal{N} \coloneqq \{1, \ldots, N\}$ and edges
$\symcal{E}$. The neighborhood of node $n\in \symcal{N}$ is defined as $\symcal{N}_n \coloneqq \{n^{\prime} \in
\symcal{N} \given \{n, n^{\prime} \} \in \symcal{E}\}$.

Associated with graph $\symcal{G}$ is the graph Laplacian matrix $\vect{L} \coloneqq \diag (\vect{W} \vect{1}_{N}) -
\vect{W}$~\cite{Bapat.graphs}, with the $N\times N$ adjacency matrix $\vect{W} = [ w_{ nn^{\prime} } ]$ defined as $w_{
  nn^{\prime} } \coloneqq 1$, if $\{ n, n^{\prime} \}\in \symcal{E}$, $w_{ nn^{\prime} } \coloneqq 0$, if $\{ n,
n^{\prime} \}\notin \symcal{E}$, and $w_{nn} \coloneqq 0$, $\forall n\in \symcal{N}$. Further, $\diag( \cdot )$
transforms a vector into a diagonal matrix with the entries of the vector placed at the main diagonal of the matrix, and
$\vect{1}_{N}$ is the $N\times 1$ all-one vector. This paper imposes no specific topology on $\symcal{G}$, unlike
federated RL, for example, which typically assumes a star topology for the network, \eg,~\cite{Zhang:2024:FedSARSA,
  Lan:2025:AFedPG}.

Agents are deployed over $\symcal{G}$, with an agent assigned to a single node. All agents share a common surrounding
environment $\{ \symcal{S}, \symcal{A}, g(\cdot) \}$, where $\symcal{S} \coloneqq \Real^{d_s}$ and $\symcal{A}$ stand
for the state and action space, respectively, for some $d_s\in \IntegerPP$, and $g(\cdot)$ is the one-step loss
function. To simplify the subsequent discussion, this manuscript considers $\symcal{A}$ to be a finite set, and even
categorical---\eg, ``move to the left'' or ``move to the right,'' as in \eqref{z.cartpole}. A state and action at agent
$n$ will be denoted henceforth by $\vect{s}_i^{(n)} \in \symcal{S}$ and $a_i^{(n)} \in \symcal{A}$, respectively, where
$i$ serves as a non-negative integer index.

For a user-defined dimensionality $d_z \in \IntegerPP$, let now also the user-defined mapping
$\vect{z}(\cdot, \cdot) \colon \symcal{S} \times \symcal{A} \to \Real^{d_z} \colon (\vect{s}, a) \mapsto \vect{z}(
\vect{s}, a)$, and the state-action space $\symcal{Z} \coloneqq \{ \vect{z} ( \vect{s}, a ) \given (\vect{s}, a) \in
\symcal{S} \times \symcal{A} \} \subset \Real^{d_z}$; in other words, $\symcal{Z}$ is the image of the mapping
$\vect{z}(\cdot, \cdot)$. Mapping $\vect{z}(\cdot,\cdot)$ is introduced to capture a broad range of state–action pairs,
including practical examples such as \eqref{z.cartpole}. Even when $\symcal{A}$ is discrete and/or categorical, the mapping $\vect{z}(\cdot, \cdot)$ is
defined so that $\symcal{Z}$ becomes a subset of the continuous space $\Real^{d_z}$, thereby enabling the subsequent use
of the kernel function $\kappa(\cdot, \cdot) \colon \symcal{Z} \times \symcal{Z} \to \Real$.  With a slight abuse of
notation, $\vect{z}$ will henceforth denote also a generic element of $\symcal{Z}$.

Agent $n$ uses its current state $\vect{s}_i^{(n)}$ and policy $\mu^{(n)} \colon \symcal{S} \to \symcal{A}$ to take
action $a_i^{(n)} \coloneqq \mu^{(n)} ( \vect{s}_i^{(n)} )$. Then, the environment provides feedback $g_i^{(n)}
\coloneqq g( \vect{z}_i^{(n)} )$ to the agent, with $\vect{z}_i^{(n)} \coloneqq \vect{z} ( \vect{s}_i^{(n)}, a_i^{(n)}
)$, via the one-step loss $g \colon \symcal{Z} \to \Real$, for the agent to transition to the new state
$\vect{s}_i^{(n)\prime} \coloneqq \vect{s}_{i+1}^{(n)}$. This transition obeys a conditional probability density
function (PDF) $p( \vect{s}_i^{(n)\prime} \given \vect{s}_i^{(n)}, a_i^{(n)} )$. This manuscript assumes that
\textit{none}\/ of the agents has \textit{any}\/ information on this conditional PDF\@.

In RL with functional approximation, Q-functions are considered to be elements of some functional space
$\symcal{H}$. The classical B-Map $T_{\diamond} \colon \symcal{H} \to \symcal{H}$ describes a ``total loss,'' comprising
the one-step loss and the expected ``minimum'' long-term loss (Q-function): $\forall Q^{(n)}\in \symcal{H}$, $\forall
\vect{z} \coloneqq \vect{z} ( \vect{s}, a ) \in \symcal{Z}$,
\begin{align}
  ( T_{\diamond} Q^{(n)} ) (\vect{z}) \coloneqq g( \vect{z} ) + \alpha \expect_{ \vect{s}^{\prime} \given \vect{z} } \{
  \inf\nolimits_{a^{\prime}\in \symcal{A} } Q^{(n)} ( \vect{z}^{\prime} ) \}\,, \label{classical.Bellman.inf}
\end{align}
where $\vect{z}^{\prime} \coloneqq \vect{z} (\vect{s}^{\prime}, a^{\prime})$, $Q^{(n)}$ and $T_{\diamond} Q^{(n)}$ are
functions defined on $\symcal{Z}$, $\alpha\in \RealPP$ is the discount factor, and $\expect_{ \vect{s}^{\prime} \given
  \vect{z} }\{ \cdot \}$ stands for conditional expectation, with $\vect{s}^{\prime}$ standing for the potential next
state after the agent takes action $a$ at state $\vect{s}$. An estimate $Q^{(n)}$ available to agent $n$ defines policy
$\mu^{(n)} \colon \symcal{S} \to \symcal{A}$ as follows~\cite{bertsekas1996neuro, bertsekas2019reinforcement}: $\forall
\vect{s} \in \symcal{S}$,
\begin{align}
  a \coloneqq \mu^{(n)}(\vect{s}) \in \arg \inf \nolimits_{ a^{\prime} \in \symcal{A} } Q^{(n)} (\, \vect{z}(\vect{s},
  a^{\prime})\, ) \,, \label{def.action}
\end{align}
where, in general, $\arg \inf$ is a set-valued operator.

It is well-known that a ``desirable'' total loss $Q_{\diamond}$, strongly connected with ``optimal policies,'' is a
fixed-point of $T_{\diamond}$---that is, $Q_{\diamond} \in \Fix T_{\diamond} \coloneqq \{ Q\in \symcal{H} \given
T_{\diamond} Q = Q \}$~\cite{bertsekas2019reinforcement}. If $\expect_{ \vect{s}^{\prime} \given \vect{z} }\{ \cdot \}$
is available, the computation of $T_{\diamond}$ and any of its fixed points can be performed at every agent $n$
independently from all other agents. Further, if the discount factor $\alpha\in (0, 1)$ and the functional space
$\symcal{H}$ is considered as the space of all (essentially) bounded functions, equipped with the sup-norm, then it can
be shown that $T_{\diamond}$ is a contraction ($\alpha$-Lipschitz continuous), which ensures that $\Fix T_{\diamond}$ is
nonempty and a singleton~\cite{bertsekas2019reinforcement, hb.plc.book}. To compute $Q_{\diamond}$, a recursive
application of~\eqref{classical.Bellman.inf} defines the classical \textit{value-iteration (VI)}\/ strategy of
RL~\cite{bertsekas2019reinforcement}.

Following the nonparametric framework of~\cite{akiyama2024nonparametric}, Q-functions are regarded here as elements of
an RKHS $\symcal{H}$~\cite{aronszajn1950, scholkopf2002learning} shared by all agents. This formulation exploits the
functional approximation capabilities of an RKHS, including Gaussian kernels, which are known to be reproducing and
possess universal-approximation properties that enable the approximation of broad classes of not necessarily continuous
functions~\cite{Gyorfi:DistrFree:10, chen96ebf}. In addition, the reproducing property of the RKHS inner product is
leveraged to allow efficient evaluation of function values via inner products.

The RKHS $\symcal{H}$ is potentially infinite dimensional~\cite{aronszajn1950, scholkopf2002learning}, equipped with a
reproducing kernel $\kappa(\cdot, \cdot) \colon \symcal{Z} \times \symcal{Z} \to \Real \colon ( \vect{z}_1, \vect{z}_2 )
\mapsto \kappa ( \vect{z}_1, \vect{z}_2 )$, so that $\kappa( \vect{z}_1, \cdot) \in \symcal{H}$, $\forall \vect{z}_1\in
\symcal{Z}$, an inner product $\innerp{\cdot}{\cdot}_{\symcal{H}}$, induced norm $\norm{\cdot}_{\symcal{H}} \coloneqq
\innerp{\cdot}{\cdot}_{\symcal{H}}^{1/2}$, and the feature mapping $\varphi \colon \symcal{Z} \to \symcal{H} \colon
\vect{z} \mapsto \varphi( \vect{z} ) \coloneqq \kappa ( \vect{z}, \cdot )$, so that the celebrated \textit{reproducing
  property}\/ holds true: $\forall Q^{(n)}\in \symcal{H}$, $\forall \vect{z}\in \symcal{Z}$, $Q^{(n)} (\vect{z}) =
\innerp{ Q^{(n)} }{ \varphi( \vect{z} ) }_{\symcal{H}}$~\cite{aronszajn1950, scholkopf2002learning}. An immediate
consequence of the reproducing property is that inner products can be computed directly through kernel function
evaluations: $\innerp{ \varphi(\vect{z}_1) }{ \varphi(\vect{z}_2) }_{\symcal{H}} = \innerp{ \kappa(\vect{z}_1, \cdot) }{
  \kappa(\vect{z}_2, \cdot) }_{\symcal{H}} = \kappa ( \vect{z}_1, \vect{z}_2)$. A well-known example of an infinite
dimensional $\symcal{H}$ is the RKHS associated with the Gaussian kernel $\kappa ( \vect{z}_1, \vect{z}_2 ) \coloneqq
\exp[ -\norm{ \vect{z}_1 - \vect{z}_2 }^2 / (2 \tau^2 ) ]$, for some $\tau\in \RealPP$.

To use familiar notations from linear algebra, the ``dot-product'' $Q^{(n)\intercal} \varphi( \vect{z} ) =
\varphi^{\intercal} ( \vect{z} ) Q^{(n)} \coloneqq \innerp{ Q^{(n)} }{ \varphi( \vect{z} ) }_{\symcal{H}}$ will be used
for the inner product hereafter, where $\intercal$ stands for the transposition operator. To simplify notation for the
distributed setting, let $\symbffrak{Q} \coloneqq [\, Q^{(1)}, \ldots, Q^{(N)}\, ] \in \symbfcal{H}$, where
$\symbfcal{H}$ stands for the $N$-times Cartesian product space $\symcal{H} \times \ldots \times \symcal{H}$, with inner
product: $\innerp{ \symbffrak{Q}_1 }{ \symbffrak{Q}_2 }_{ \symbfcal{H} } \coloneqq \sum_{n=1}^N \innerp{ Q_1^{(n)} }{
  Q_2^{(n)} }_{\symcal{H}}$, $\forall \symbffrak{Q}_1, \symbffrak{Q}_2\in \symbfcal{H}$. For convenience, square
brackets $[ \cdot ]$ are used instead of parentheses $( \cdot )$ to denote tuples $\symbffrak{Q}$ in $\symbfcal{H}$, and
thus allowing for the use of familiar linear algebra operations in the following sections.

\subsection{General assumptions for the decentralized setting}\label{sec:setting}

The following assumptions will serve as overarching assumptions throughout the discussion.

\begin{assumptions}\label{ass:setting}
  \mbox{}
  \begin{assslist}

  \item\label{ass:setting.topology} \textbf{(Graph topology)} The graph $\symcal{G} \coloneqq (\symcal{N}, \symcal{E})$
    is undirected and connected~\cite{Gross.GraphTheory.book}, and its topology can be \textit{arbitrary.} Each agent
    $n$ exchanges information only with its immediate neighbors $\symcal{N}(n)$, and no centralized node capable of
    gathering all data and performing computations is available.

  \item\label{ass:setting.environment} \textbf{(Common environment)} All agents interact with a shared environment
    $(\symcal{S}, \symcal{A}, g)$, where the state space $\symcal{S}$ is continuous, and the action space $\symcal{A}$
    is discrete---even categorical---with finite cardinality.

  \item\label{ass:setting.trajectory} \textbf{(Trajectory data)} Agents lack statistical knowledge required to compute
    $\expect_{ \vect{s}^{\prime} \given \vect{z} }\{ \cdot \}$ in~\eqref{classical.Bellman.inf}. Instead, each agent $n$
    relies exclusively on its own private trajectory data:
    \begin{align*}
      \symcal{T}^{(n)} \coloneqq \{\, (\vect{s}_i^{(n)}, a_i^{(n)}, g_i^{(n)}, \vect{s}_i^{(n)\prime} =
      \vect{s}_{i+1}^{(n)} )\, \}_{i=1}^{ N_{\textnormal{av}}^{(n)} } \,,
    \end{align*}
    for some positive integer $N_{\textnormal{av}}^{(n)}$. All computations are performed in batch mode; no online
    processing is considered. Datasets $\symcal{T}^{(n)}$ and $\symcal{T}^{(n^{\prime})}$ need not be identical, or
    even partially overlapping, $\forall n, n^{\prime} \in \symcal{N}$.

  \item\label{ass:setting.share.Q} \textbf{(Q-function sharing)} Each agent $n$ shares a copy of its Q-function estimate
    $Q^{(n)}$ with its neighbors $\symcal{N}_n$.

  \end{assslist}
\end{assumptions}

As per \cref{ass:setting}, agents must rely on their private trajectory data and cooperation to collectively approximate
a fixed point of $T_{\diamond}$. In decentralized fitted Q-iteration~\cite{distributed.TD}, for example, sequence $(\,
\symbffrak{Q}[k] \coloneqq [\, Q^{(1)}[k], \ldots, Q^{(N)}[k]\, ] \,)_{ k\in \IntegerP }$ is generated according to
VI~\cite{bertsekas2019reinforcement}, with the non-negative integer $k$ being the VI index:
\begin{align}
  \symbffrak{Q} [k+1] \coloneqq \vect{T}_{\textnormal{TD}}( \symbffrak{Q} [k] ) \,, \label{D-FQ}
\end{align}
so that $( \symbffrak{Q} [k] )_{ k\in \IntegerP }$ converges, as $k\to \infty$, to a fixed point of the consensus-based
Bellman mapping $\vect{T}_{\textnormal{TD}} \colon \symbfcal{H} \to \symcal{C}_{ \symbfcal{H} } \colon \symbffrak{Q}
\mapsto \vect{T}_{\textnormal{TD}}( \symbffrak{Q} )$, defined as
\begin{align}
  \vect{T}_{\textnormal{TD}}( \symbffrak{Q} ) & \in \arg \min\nolimits_{ \symbffrak{Q}^{\prime} \in \symcal{C}_{
      \symbfcal{H} } } \sum\nolimits_{ n\in\symcal{N} } \symcal{L}_{ \textnormal{TD} }^{(n)} ( Q^{(n)\prime};\, Q^{(n)}
  ) \,, \label{Bellman.TD}
\end{align}
where the classical temporal-difference (TD) loss
\begin{align*}
  & \symcal{L}_{ \textnormal{TD} }^{(n)} ( Q^{(n) \prime};\, Q^{(n)} ) \\
  & \coloneqq \tfrac{1}{2} \sum_{i=1}^{ N_{\textnormal{av}}^{(n)} } \Bigl[ g_i^{(n)} + \alpha\, \inf_{ \mathclap{
        a_i^{(n)\prime} \in \symcal{A} } }\, Q^{(n)} (\vect{s}_i^{(n)\prime}, a_i^{(n)\prime }) -
    Q^{(n)\prime}(\vect{z}_i^{(n)}) \Bigr]^2 \,,
\end{align*}
and the \textit{consensus set}\/
\begin{alignat}{2}
  \symcal{C}_{ \symbfcal{H} }
  & {} \coloneqq {}
  && \{ \symbffrak{Q}^{\prime} \in \symbfcal{H} \given Q^{(n) \prime} = Q^{(n^{\prime}) \prime}, \forall \{n,
  n^{\prime}\} \in \symcal{E} \} \notag\\
  & = && \{ \symbffrak{Q}^{\prime} \in \symbfcal{H} \given Q^{(1) \prime} = \ldots = Q^{(N) \prime} \}
         \,, \label{consensus.H}
\end{alignat}
with the latter expression of $\symcal{C}_{ \symbfcal{H} }$ in \eqref{consensus.H} following from
\cref{ass:setting.topology} that $\symcal{G}$ is connected~\cite{Gross.GraphTheory.book}.

As per \cref{ass:setting}, no single node can compute~\eqref{Bellman.TD}, necessitating a distributed solution. The work
of~\cite{distributed.LSTD.ADMM} adopts the same setting as~\cite{distributed.TD}, employs a least-squares (LS)TD-type
loss, and addresses the resulting problem via an \textit{inexact}\/ variant of the popular alternating direction method
of multipliers (ADMM)~\cite{boyd2011distributed}. Studies~\cite{distributed.A3C, distributed.deep} assume a star
topology for $\symcal{G}$. Moreover, \cite{distributed.LSTD, distributed.ActorCritic, distributed.LSTD.primal,
  distributed.Gossip, distributed.A3C, distributed.deep} focus on streaming data and operate in online-learning or
stochastic-optimization modes. To broaden the set of benchmarks for evaluating the proposed \cref{algo}, this work also
introduces in \cref{sec:numerical} a distributed method for solving~\eqref{Bellman.TD} via the \textit{exact}\/ version
of ADMM.

\section{The Centralized Bellman Mapping}\label{sec:central.Bmap}

Had there been a \textit{centralized node}\/ $n_{\star} \in \symcal{N}$, connected with every node of the graph
$\symcal{G}_{\star} = ( \symcal{N}, \symcal{E}_{\star} )$, where $\symcal{E}_{\star}$ follows a \textit{star topology,}
able to \textit{collect all}\/ data $\{ \symcal{T}^{(n)} \}_{ n\in \symcal{N} }$ of \cref{ass:setting.trajectory} and to
perform all necessary computations, a centralized B-Map $T_{\odot} \colon \symcal{H} \to \symcal{H} \colon Q \mapsto
T_{\odot} ( Q )$ could have been defined at $n_{\star}$ as
 \begin{alignat}{2}
   T_{\odot} ( Q )
   & {} \coloneqq {} && \sum_{ n\in\symcal{N} } \sum_{i=1}^{ N_{\textnormal{av}}^{(n)} } \overbrace{ [\,
       g_i^{(n)} + \alpha \inf_{ a_i^{(n) \prime} \in \symcal{A} } Q (\, \vect{z}( \vect{s}_i^{(n)\prime},
       a_i^{(n)\prime}) \, ) \,] }^{ c_i^{(n)}(Q) } \, \psi_{\odot i}^{(n)} \notag\\
   & \coloneqq && \sum_{ n\in\symcal{N} } \vect{\Psi}_{\odot}^{(n)}\, \vect{c}^{(n)}( Q ) \,, \label{Bmap.star}
   \intertext{where}
   g_i^{(n)} & \coloneqq && g( \vect{z}_i^{(n)} )\,, \quad  \vect{z}_i^{(n)} \coloneqq \vect{z} ( \vect{s}_i^{(n)},
   a_i^{(n)} )\,, \notag \\
   \vect{c}^{(n)}( Q )
   & \coloneqq && [ c_1^{(n)}( Q ), \ldots, c_{ N_{\textnormal{av}}^{(n)} }^{(n)}( Q ) ]^{\intercal} \,, \notag \\
   \vect{\Psi}_{\odot}^{(n)}
   & \coloneqq && [ \psi_{\odot 1}^{(n)}, \ldots, \psi_{ \odot N_{\textnormal{av}}^{(n)} }^{(n)} ] \,, \notag
\end{alignat}
and $\{ \{ \psi_{\odot i}^{(n)} \}_{ i=1 }^{ N_{\textnormal{av}}^{(n)} } \}_{ n \in \symcal{N} }$ are user-defined
\textit{basis}\/ functions/elements drawn from the RKHS $\symcal{H}$, which is endowed with a reproducing kernel
$\kappa$, a feature mapping $\varphi$, and an inner product $\innerp{\cdot}{\cdot}_{\symcal{H}}$, consistent with the
discussion at the end of \cref{sec:preliminaries}.

Form \eqref{Bmap.star} is inspired by the non-distributed design of~\cite[(3b)]{akiyama2024nonparametric}, which was
introduced as a surrogate for the classical formulation~\eqref{classical.Bellman.inf} in settings where computing the
conditional expectation in~\eqref{classical.Bellman.inf} is infeasible; see \cref{ass:setting.trajectory}. In
\eqref{Bmap.star}, this conditional expectation is approximated by a linear combination of user-defined functions $\{
\psi_{\odot i}^{(n)} \} \subset \symcal{H}$, with coefficients $c_i^{(n)} (Q^{(n)})$ determined by evaluations of the
one-step cost $g$ and the long-term $Q$-functions at $\{ \symcal{T}^{(n)} \}_{n \in \symcal{N}}$.

Study~\cite[Prop.~1]{akiyama2024nonparametric} develops a variational framework for designing $\{ \psi_{\odot i}^{(n)}
\}$. Remarkably, by selecting appropriate loss functions and regularization terms within that variational framework, one
recovers several well-known B-Map designs~\cite[Prop.~1]{akiyama2024nonparametric}. This work, for the sake of clarity
and concreteness, adopts the following specific basis functions:%
\begin{subequations}\label{proposed.basis.functions.central}
  \begin{alignat}{2}
    \vect{\Psi}_{\odot}^{(n)}
    & {} \coloneqq {}
    && (\, \sum\nolimits_{n\in\symcal{N}} \vect{\Phi}^{(n)} \vect{\Phi}^{(n)\intercal} + \sigma \Id\, )^{-1}\,
    \vect{\Phi}^{(n)} \\
    & = && (\, \vect{\Phi}_{ \symcal{N} } \vect{\Phi}_{ \symcal{N} }^{\intercal} + \sigma \Id\, )^{-1}\,
    \vect{\Phi}^{(n)} \,, \label{psi.star.n}
    \intertext{where}
           \vect{\Phi}^{(n)}
    & {} \coloneqq {} && [ \varphi(\vect{z}_1^{(n)}), \ldots, \varphi( \vect{z}_{ N_{\textnormal{av}}^{(n)} }^{(n)})
           ]\,, \notag \\
    \vect{\Phi}^{(n)} \vect{\Phi}^{(n)\intercal}
    & = && \sum\nolimits_{i = 1}^{ N_{\textnormal{av}}^{(n)} } \varphi(\vect{z}_i^{(n)}) \varphi^{\intercal}
    (\vect{z}_i^{(n)}) \,, \label{Cov.Matrix.nodal} \\
    \vect{\Phi}_{ \symcal{N} }
    & \coloneqq && [ \vect{\Phi}^{(1)}, \ldots, \vect{\Phi}^{(N)} ] \,, \notag \\
    \vect{\Phi}_{ \symcal{N} } \vect{\Phi}_{ \symcal{N} }^{\intercal}
    & = && \sum\nolimits_{n\in\symcal{N}} \vect{\Phi}^{(n)} \vect{\Phi}^{(n)\intercal} \,, \label{Cov.Matrix.global}
  \end{alignat}%
\end{subequations}%
$\Id \colon \symcal{H} \to \symcal{H}$ is the identity operator, and $\sigma \in \RealPP$. Borrowing from the
signal-processing jargon, \eqref{Cov.Matrix.nodal} will be called the \textit{nodal covariance operator,}
while~\eqref{Cov.Matrix.global} the \textit{network-wide covariance operator.} Notice that~\eqref{Cov.Matrix.nodal}
and~\eqref{Cov.Matrix.global} operate in the feature space $\symcal{H}$.

A certain degree of design flexibility is available, as any choice of $\vect{\Psi}_{\odot}^{(n)}$ from the framework of
\cite[Prop.~1]{akiyama2024nonparametric}, other than \eqref{proposed.basis.functions.central}, could in principle be
employed. However, \eqref{psi.star.n} is adopted here because it has a simpler form than the other alternatives offered
in \cite[Prop.~1]{akiyama2024nonparametric}, enables a direct extension of the design in \cite{akiyama2024nonparametric}
to the current distributed setting, and possesses rigorously established theoretical
properties~\cite[Sec.~II.D]{akiyama2024nonparametric}.

Under the specific \eqref{psi.star.n}, \eqref{Bmap.star} takes the following special form:%
\begin{subequations}\label{Bmap.star.special}%
  \begin{alignat}{2}
    T_{\odot} ( Q )
    & = && \sum\nolimits_{n\in \symcal{N}} \vect{\Psi}_{\odot}^{(n)}\, \vect{c}^{(n)}( Q ) \label{Bmap.star.special.0} \\
    & = && \sum\nolimits_{n\in \symcal{N}} (\, \vect{\Phi}_{ \symcal{N} } \vect{\Phi}_{ \symcal{N} }^{\intercal} +
    \sigma \Id\, )^{-1} \vect{\Phi}^{(n)}\, \vect{c}^{(n)}( Q ) \label{Bmap.star.special.i} \\
    % & = (\, \vect{\Phi}_{ \symcal{N} } \vect{\Phi}_{ \symcal{N} }^{\intercal} + \sigma \Id\,
    % )^{-1} \vect{\Phi}_{ \symcal{N} } \begin{bmatrix}
    %   \vect{c}^{(1)}( Q ) \\ \rvdots \\
    %   \vect{c}^{(N)}( Q )
    %   \end{bmatrix} \notag \\
    & = && (\, \vect{\Phi}_{ \symcal{N} } \vect{\Phi}_{ \symcal{N} }^{\intercal} + \sigma \Id\, )^{-1}
    \vect{\Phi}_{ \symcal{N} } \, \vect{c}_{ \symcal{N} }( Q ) \label{Bmap.star.special.ii} \\
    & = && \vect{\Phi}_{ \symcal{N} } (\, \vect{K}_{ \symcal{N} } + \sigma \Id\, )^{-1}\, \vect{c}_{ \symcal{N} }( Q )
    \,, \label{Bmap.star.special.iii}
    \intertext{where}
    \vect{c}_{ \symcal{N} }( Q )
    & {} \coloneqq {} && [\, \vect{c}^{(1)\intercal} ( Q ), \ldots, \vect{c}^{(N)\intercal} ( Q ) \, ]^{\intercal}
    \notag\\
    \vect{K}_{ \symcal{N} }
    & \coloneqq && \vect{\Phi}_{ \symcal{N} }^{\intercal} \vect{\Phi}_{ \symcal{N} } \,, \notag
  \end{alignat}%
\end{subequations}%
and the equality in \eqref{Bmap.star.special.iii} can be easily verified.

In the presence of a centralized node $n_{\star} \in \symcal{N}$, the computation of the centralized $T_{\odot}$
in~\eqref{Bmap.star.special} and its fixed point, if it exists, is immediate. More precisely, the following two-step
algorithmic procedure suffices to compute and distribute a fixed point $Q_{ \odot}$ of $T_{\odot} (\cdot)$ to all nodes
across the graph.

\begin{algo}[Centralized solution]\label{algo.central}\mbox{}

  \begin{algorithmic}[1]

    \Require Graph $\symcal{G}_{\star} = ( \symcal{N}, \symcal{E}_{\star} )$ with a centralized node $n_{\star} \in
    \symcal{N}$ ($\symcal{E}_{\star}$ follows a star topology), data $\{ \symcal{T}^{(n)} \}_{ n\in \symcal{N} }$,
    reproducing kernel $\kappa$.

    \State{Each node $n$ sends its data $\symcal{T}^{(n)}$ to the centralized node $n_{\star}$.}

    \State{With all data $\{ \symcal{T}^{(n)} \}_{ n\in \symcal{N} }$ available, the centralized node computes
      $T_{\odot} (\cdot)$ via~\eqref{Bmap.star.special}, identifies a fixed point $Q_{\odot} \in \Fix T_{\odot}$, and
      distributes it to all nodes across the graph, so that actions per node are taken according to the ``optimal''
      policy $\mu_{\odot} \colon \symcal{S} \to \symcal{A} \colon \vect{s} \mapsto \mu_{\odot} (\vect{s}) \in \arg \inf
      \nolimits_{ a^{\prime} \in \symcal{A} } Q_{ \odot} ( \vect{s}, a^{\prime}
      )$.}\label{algo:compute.at.fusion.center}

  \end{algorithmic}

\end{algo}

\section{A Novel Distributed Value-Iteration Algorithm}\label{sec:proposed.solution}

\subsection{Challenges in the absence of a centralized node}\label{sec:challenges}

In the absence of a centralized node, \cref{algo.central} is infeasible. This section proposes a fully distributed
alternative which abides by \cref{ass:setting}.

To this end, define first the following \textit{nodal}\/ B-Maps: $T^{(n)}\colon \symcal{H} \to \symcal{H} \colon Q
\mapsto T^{(n)}(Q)$ with
\begin{align}
  T^{(n)}(Q)
  & \coloneqq \sum_{i=1}^{ N_{\textnormal{av}}^{(n)} } [\, g_i^{(n)} + \alpha \inf_{ a_i^{(n) \prime} \in \symcal{A}  }
    Q ( \vect{s}_i^{(n)\prime}, a_i^{(n)\prime}) \,] \, \psi_i^{(n)} \notag \\
  & = \vect{\Psi}^{(n)}\, \vect{c}^{(n)}( Q ) \,, \label{Bmap.nodal}
\end{align}
where $\vect{\Psi}^{(n)} \coloneqq [ \psi_1^{(n)}, \ldots, \psi_{ N_{\textnormal{av}}^{(n)} }^{(n)} ]$ and $\{
\psi_{i}^{(n)} \}_{ i=1 }^{ N_{\textnormal{av}}^{(n)} }$ are basis functions in $\symcal{H}$ defined by the agent at
node $n$. Following the structure of \eqref{psi.star.n}, a natural choice for $\vect{\Psi}^{(n)}$ is
\begin{align}
  \vect{\Psi}^{(n)} = (\, \vect{C}^{(n)} + \sigma \Id \, )^{-1}\, \vect{\Phi}^{(n)} \,, \label{psi.n.k.guess}
\end{align}
where $\vect{C}^{(n)}$ is a nodal estimate of the network-wide covariance operator $\vect{\Phi}_{ \symcal{N} }
\vect{\Phi}_{ \symcal{N} }^{\intercal}$; hence, \eqref{psi.n.k.guess} serves as an estimate of \eqref{psi.star.n}. A
more detailed form appears in \eqref{psi.n.k}. However, such a design raises the following issues.

\begin{challenges}\label{all.challenges}\mbox{}
  \begin{challengelist}%

  \item\label{challenge1} \textbf{(Distribute Q-functions)} A consensus-based distributed algorithm over $\symcal{G}$ is
    needed to approximate the centralized computation of $\sum_{n^{\prime} \in\symcal{N}} \vect{\Psi}_{\odot}^{(
      n^{\prime} )}\, \vect{c}^{( n^{\prime} )}(\cdot)$ in~\eqref{Bmap.star.special.0} at every node $n$.

  \item\label{challenge2} \textbf{(Distribute covariance operators)} A consensus-based distributed algorithm over
    $\symcal{G}$ is required to ensure that the estimate $\vect{C}^{(n)}$ in~\eqref{psi.n.k.guess} accurately
    approximates the network-wide covariance operator~\eqref{Cov.Matrix.global} at every node $n$.

  \item\label{challenge3} \textbf{(Curse of dimensionality)} Because $\symcal{H}$ may be infinite-dimensional,
    addressing \cref{challenge1,challenge2} involves sharing high- or even infinite-dimensional objects, such as
    Q-functions and covariance operators. To mitigate the resulting communication bandwidth constraints, a
    dimensionality-reduction scheme is required.

  \end{challengelist}%
\end{challenges}

\subsection{Distributing Q-functions}\label{sec:distribute.Q}

To address \cref{challenge1}, gather first all nodal B-Maps into the following \textit{in-network}\/ B-Map: $\vect{T}
\colon \symbfcal{H} \to \symbfcal{H} \colon \symbffrak{Q} = [ Q^{(1)}, \ldots, Q^{(N)} ] \mapsto \vect{T} (
\symbffrak{Q} )$ with
\begin{align}
  \vect{T} ( \symbffrak{Q} ) \coloneqq [\, T^{(1)}( Q^{(1)} ),
  \ldots, T^{(N)}( Q^{(N)} )\, ] \,. \label{Bmap.in.net}
\end{align}
This paper's counter-proposition to \eqref{Bellman.TD} is to solve distributively the following task:%
\begin{subequations}%
  \begin{align}
    & \arg \min_{ \symbffrak{Q}^{\prime} \in \symcal{C}_{\symbfcal{H}} } \tfrac{1}{2} \norm{\, \symbffrak{Q}^{\prime} -
      N\, \vect{T}( \symbffrak{Q} )\, }_{ \symbfcal{H} }^2 \label{Q.task} \\
    & = \arg \min_{ \symbffrak{Q}^{\prime} \in \symcal{C}_{\symbfcal{H}} } \sum\nolimits_{n\in\symcal{N}} \tfrac{1}{2}
    \norm{\, Q^{\prime (n)} - N\, T^{(n)} ( Q^{(n)} )\, }_{\symcal{H}}^2 \notag\\
    & = \left [\, \sum\nolimits_{n\in\symcal{N}} T^{(n)} ( Q^{(n)} ), \ldots, \sum\nolimits_{n\in\symcal{N}}
      T^{(n)} ( Q^{(n)} ) \, \right ] \label{solution.Q.task} \\
    & = [\, \vect{T}(\symbffrak{Q})\, \vect{1}_N, \ldots, \vect{T}(\symbffrak{Q})\, \vect{1}_N\, ] \,. \notag
  \end{align}
\end{subequations}%
The closed-form solution \eqref{solution.Q.task} of \eqref{Q.task} is straightforward for a centralized node, but in its
absence, a distributed algorithmic approach is required. To this end, the framework of~\cite{slavakis2018fejer} is
adopted due to its generality, flexibility, and simple recursive structure. Accordingly, the linear operators $A^{
  \textnormal{Q} }, A_{\varpi}^{ \textnormal{Q} } \colon \symbfcal{H} \to \symbfcal{H}$ are defined by%
\begin{subequations}\label{Aq.maps}%
  \begin{align}
    A^{ \textnormal{Q} }( \symbffrak{Q} )
    & \coloneqq \symbffrak{Q}( \vect{I}_N - \gamma \vect{L} )\,, \label{operator.Aq}\\
    A_{\varpi}^{ \textnormal{Q} } ( \symbffrak{Q} )
    & \coloneqq \varpi A^{ \textnormal{Q} } ( \symbffrak{Q} ) + (1-\varpi) \symbffrak{Q}\,, \label{operator.Aq.beta}
  \end{align}
\end{subequations}
$\forall \symbffrak{Q}\in \symbfcal{H}$. In~\eqref{Aq.maps}, $\varpi \in [1/2, 1)$ and $\gamma \in (\, 0, 1 /
  \norm{\vect{L}}_2 \, ]$, where $\norm{ \vect{L} }_2$ is the spectral norm of the Laplacian matrix
$\vect{L}$~\cite{Ben-Israel:03}.

Owing to the definition of the Laplacian matrix, notice that the $n$th entry of $A^{ \textnormal{Q} } (\symbffrak{Q})$
in~\eqref{operator.Aq} takes the following form:
\begin{align}
  (\, A^{ \textnormal{Q} } (\symbffrak{Q})\, )^{(n)} = (1 -
  \gamma \lvert \symcal{N}_n \rvert)\, Q^{(n)} + \gamma
  \sum_{ n^{\prime} \in \symcal{N}_n } Q^{(
  n^{\prime} )} \,. \label{Aq.n}
\end{align}
This is a clear demonstration of the distributive nature of $A^{ \textnormal{Q} }$, because not only the local
Q-function $Q^{(n)}$ but also copies of $\{ Q^{( n^{\prime} )} \}_{ n^{\prime} \in \symcal{N}_n }$ need to be
transmitted from neighbors $\symcal{N}_n$ to node $n$ to compute $(\, A^{ \textnormal{Q} } (\symbffrak{Q})\,
)^{(n)}$. Because of $\gamma \in (\, 0, 1 / \norm{\vect{L}}_2 \, ]$, it can be verified that $\norm{ A^{ \textnormal{Q}
  } } = \norm{ \vect{I}_N - \gamma \vect{L} }_2 \leq 1$, where $\norm{ A^{ \textnormal{Q} } }$ is the norm induced by
  the inner product $\innerp{\cdot}{\cdot}_{ \symbfcal{H} }$. Moreover, because $\gamma \leq 1 / \norm{ \vect{L} }_2$,
  it can be verified that $\innerp{ A^{ \textnormal{Q} } (\symbffrak{Q}) }{ \symbffrak{Q} }_{ \symbfcal{H} } \geq 0$,
  $\forall \symbffrak{Q}\in \symbfcal{H}$, that is, $A^{ \textnormal{Q} }$ is positive~\cite[\S9.3]{kreyszig:91}. Now,
  notice that $\vect{L} \vect{1}_N = \vect{0} = 0\cdot \vect{1}_N$. By \cref{ass:setting.topology} and the fact that
  $\vect{L}$ is positive semidefinite~\cite[Lem.~4.3]{Bapat.graphs}, the rank of $\vect{L}$ is $N-1$ with $0$ being its
  smallest eigenvalue, and the kernel space $\kernel \vect{L} = \Span( \vect{1}_N )$, so that
\begin{align}
  \Fix ( A^{ \textnormal{Q} } ) \coloneqq \Set{ \symbffrak{Q}\in \symbfcal{H} \given A^{ \textnormal{Q} } (
    \symbffrak{Q} ) = \symbffrak{Q} } = \symcal{C}_{\symbfcal{H}} \,. \label{Fix.Aq}
\end{align}

As in~\eqref{D-FQ}, $k\in \IntegerP$ serves as the VI index in this paper; see \cref{fig:algo}. With $\symbffrak{Q}[k] =
[ Q^{(1)}[k], \ldots, Q^{(N)}[k]\, ]\in \symbfcal{H}$ being the snapshot of all Q-function estimates across $\symcal{G}$
at VI iteration $k$, the aforementioned properties of $A^{ \textnormal{Q} }$ and~\eqref{Fix.Aq} ensure that the sequence
$( \symbffrak{Q}_m[k] )_{m\in \IntegerP}$ generated by $\symbffrak{Q}_{-1}[k] \coloneqq [ 0, \ldots, 0 ]$ and, $\forall
m\in \IntegerP$,%
\begin{subequations}\label{ACFB.Q}%
  \begin{alignat}{2}
    \symbffrak{Q}_0[k]
    & {} \coloneqq {}
    && A_{\varpi}^{ \textnormal{Q} } (\symbffrak{Q}_{-1}[k]) - \eta (\, \symbffrak{Q}_{-1}[k] - N\, \vect{T}(
    \symbffrak{Q} [k] )\, ) \,, \label{ACFB.Q.zero} \\
    \symbffrak{Q}_{m+1}[k]
    & \coloneqq
    && \symbffrak{Q}_{m}[k] - (\, A_{\varpi}^{ \textnormal{Q} } (\symbffrak{Q}_{m-1}[k]) - \eta\,
    \symbffrak{Q}_{m-1}[k]\, ) \notag \\
    &&& \phantom{\symbffrak{Q}_{m}[k]} + (\, A^{ \textnormal{Q} } (\symbffrak{Q}_{m}[k]) - \eta\, \symbffrak{Q}_{m}[k]
        \, ) \,, \label{ACFB.Q.m}
  \end{alignat}
\end{subequations}
with $\eta\in ( 0, 2(1-\varpi)\, )$, converges strongly (recall that $\symcal{H}$ may be infinite dimensional) to the
solution~\eqref{solution.Q.task} as $m\to \infty$~\cite[Lem.~3.4 and Cor.~3.5]{slavakis2018fejer}. Even more, a linear
convergence rate for $( \symbffrak{Q}_m[k] )_{m\in \IntegerP}$ is established by \cref{thm:Q.consensus.rate}.

\cref{challenge3} appears prominently in the previous discussion, because possibly infinite dimensional Q-functions need
to be shared among neighbors to compute~\eqref{Aq.n}. To surmount \cref{challenge3}, dimensionality reduction is
needed. To this end, the feature mapping $\varphi \colon \symcal{Z} \to \symcal{H}$---recall the discussion at the end
of \cref{sec:preliminaries}---will be replaced henceforth by the random-Fourier-feature-(RFF) mapping~\cite{rff}
$\tilde{\varphi} \colon \symcal{Z} \to \Real^D \colon \vect{z} \mapsto \tilde{\varphi}( \vect{z} )$, for a user-defined
$D\in \IntegerPP$, with
\begin{align}
  \tilde{\varphi}( \vect{z} ) \coloneqq \sqrt{ \tfrac{2}{D} }\, [ \cos (\vect{v}_1^{\intercal} \vect{z} + u_1), \ldots,
  \cos (\vect{v}_{D}^{\intercal} \vect{z} + u_{D})]^{\intercal}  \,, \label{RFF}
\end{align}
$\forall \vect{z}\in \symcal{Z}$, where $\{ \vect{v}_i \}_{i=1}^D$ and $\{ u_i \}_{i=1}^D$ are samples from the Gaussian
and uniform distributions, respectively. In other words, a general Q-function in the potentially infinite-dimensional
$\symcal{H}$, for instance $Q^{(n)} = \sum_i c_i^{(n)} \varphi( \vect{z}_i^{(n)} )$, will hereafter be represented by
the dimensionally reduced $D \times 1$ vector $\tilde{Q}^{(n)} = \sum_i c_i^{(n)} \tilde{\varphi} ( \vect{z}_i^{(n)}
)$. Although $\tilde{\varphi}$ formally replaces $\varphi$, the symbol $\varphi$ will continue to be used for clarity
and notational simplicity, with the understanding that $\tilde{\varphi}$ is implemented in the
background.

\subsection{Distributing covariance operators}\label{sec:distribute.cov}

Drawing now attention to \cref{challenge2}, after the RFF dimensionality-reduction scheme has been applied, a
distributed scheme is needed to compute the $D \times D$ network-wide covariance matrix $\vect{\Phi}_{ \symcal{N} }
\vect{\Phi}_{ \symcal{N} }^{\intercal}$ of~\eqref{Cov.Matrix.global}; recall that the RFF $\tilde{\varphi}$ is
implemented now in computations. To this end, define the linear vector space of operators $\symbfcal{O} \coloneqq
\Real^{D \times ND} = \{ \symbffrak{C} \coloneqq [ \vect{C}^{(1)}, \ldots, \vect{C}^{(N)} ] \given \vect{C}^{(n)}\in
\Real^{D\times D} \}$, equipped with the standard inner product $\innerp{ \symbffrak{C}_1 }{ \symbffrak{C}_2 }_{
  \symbfcal{O} } \coloneqq \trace( \symbffrak{C}_1^{\intercal} \symbffrak{C}_2 )$, $\forall \symbffrak{C}_1,
\symbffrak{C}_2\in \symbfcal{O}$. Observe then%
\begin{subequations}
  \begin{align}
    & \arg\min_{ \symbffrak{C} \in
      \symcal{C}_{ \symbfcal{O} } }
      \tfrac{1}{2}  \norm{\,
      \symbffrak{C} - N\, \symbffrak{C}_{ \symcal{N} }\,}_{
      \textnormal{F} }^2 \label{P.task} \\
    & = \arg\min_{ \symbffrak{C} \in
      \symcal{C}_{ \symbfcal{O} } }
      \sum\nolimits_{n\in\symcal{N}}  \tfrac{1}{2}
      \norm{\, \vect{C}^{(n)} - N\, \vect{\Phi}^{(n)}
      \vect{\Phi}^{(n)\intercal} \,}_{ \textnormal{F} }^2 \notag \\
      & = [ \vect{\Phi}_{ \symcal{N} } \vect{\Phi}_{
      \symcal{N} }^{\intercal}, \ldots, \vect{\Phi}_{
      \symcal{N} } \vect{\Phi}_{ \symcal{N}
      }^{\intercal} ] \,, \label{solution.P.task}
  \end{align}
\end{subequations}
where $\norm{ \cdot }_{ \textnormal{F} }$ is the Frobenius norm,
\begin{align}
  \symbffrak{C}_{ \symcal{N} } \coloneqq [
  \vect{\Phi}^{(1)} \vect{\Phi}^{(1)\intercal}, \ldots,
  \vect{\Phi}^{(N)} \vect{\Phi}^{(N)\intercal}] \label{C.nodes}
\end{align}
gathers all nodal covariance operators, and the consensus set
\begin{alignat*}{2}
  \symcal{C}_{ \symbfcal{O} }
  & {} \coloneqq {}
  && \{ \symbffrak{C} \in \symbfcal{O} \given \vect{C}^{(n)} =
     \vect{C}^{(n^{\prime})}, \forall \{n,
     n^{\prime}\} \in \symcal{E} \} \\
  & = && \{ \symbffrak{C} \in \symbfcal{O} \given
         \vect{C}^{(1)} = \ldots = \vect{C}^{(N)} \} \,.
\end{alignat*}
Along the lines of~\eqref{Aq.maps}, define the linear operators $A^{ \textnormal{C} }, A_{\varpi}^{ \textnormal{C} }
\colon \symbfcal{O} \to \symbfcal{O}$ by%
\begin{subequations}\label{Ap.maps}%
  \begin{align}
    A^{ \textnormal{C} } ( \symbffrak{C} )
    & \coloneqq \symbffrak{C}\, (\, ( \vect{I}_N - \gamma
      \vect{L} ) \otimes \vect{I}_D\, )
      \,, \label{operator.Ap}\\
    A_{\varpi}^{ \textnormal{C} } ( \symbffrak{C} )
    & \coloneqq \varpi A^{ \textnormal{C} } ( \symbffrak{C} ) +
      (1 - \varpi) \symbffrak{C} \,, \label{operator.Ap.beta}
  \end{align}
\end{subequations}
$\forall \symbffrak{C}\in \symbfcal{O}$, where $\otimes$ stands for the Kronecker product, $\gamma \in (\, 0, 1 /
\norm{\vect{L}}_2 \, ]$, and $\varpi \in [1/2, 1)$. It is not difficult to verify that per node $n$, only
    $\vect{C}^{(n)}$ and copies of $\{ \vect{C}^{( n^{\prime} )} \}_{n^{\prime} \in \symcal{N}_n}$ from the neighboring
    agents need to be shared to compute
\begin{align}
  (\, A^{ \textnormal{C} } ( \symbffrak{C})\, )^{(n)} = (1 -
  \gamma \lvert \symcal{N}_n \rvert)\, \vect{C}^{(n)} + \gamma
  \sum_{ n^{\prime} \in \symcal{N}_n } \vect{C}^{(
  n^{\prime} )} \,. \label{Ap.n}
\end{align}
Moreover, similarly to the discussion following~\eqref{Aq.n} and by using basic properties of $\otimes$, it can be
verified that $\norm{ A^{ \textnormal{C} } } = \norm{ ( \vect{I}_N - \gamma \vect{L} ) \otimes \vect{I}_D }_2 = \norm{
  \vect{I}_N - \gamma \vect{L} }_2 \leq 1$, that $A^{ \textnormal{C} }$ is positive, that $( \vect{L} \otimes \vect{I}_D
) ( \vect{1}_N \otimes \vect{I}_D ) = ( \vect{L} \vect{1}_N ) \otimes \vect{I}_{D} = \vect{0}$, that $\kernel ( \vect{L}
\otimes \vect{I}_D ) = \Span( \vect{1}_N \otimes \vect{I}_D )$, and that
\begin{align}
  \Fix ( A^{ \textnormal{C} } ) \coloneqq \Set{ \symbffrak{C}
  \in \symbfcal{O} \given A^{ \textnormal{C} } (
  \symbffrak{C} ) = \symbffrak{C} } = \symcal{C}_{
  \symbfcal{O} } \,. \label{Fix.Ap}
\end{align}
Consequently, and similarly to~\eqref{ACFB.Q}, sequence
$(\, \symbffrak{C}_l = ( \vect{C}_l^{(1)}, \ldots, \vect{C}_l^{(N)} )\, )_{l\in \IntegerP}$
generated by $\symbffrak{C}_{-1} \coloneqq ( \vect{0}, \ldots, \vect{0} )$ and%
\begin{subequations}\label{ACFB.C}%
  \begin{alignat}{2}
    \symbffrak{C}_0
    & {} \coloneqq {}
    && A_{\varpi}^{ \textnormal{C} } (\symbffrak{C}_{-1}) -
       \eta ( \symbffrak{C}_{-1} - N\,
       \symbffrak{C}_{ \symcal{N} } ) \,, \label{ACFB.C.zero} \\
    \symbffrak{C}_{l+1}
    & \coloneqq
    && \symbffrak{C}_{l} - (\, A_{\varpi}^{ \textnormal{C} }
       (\symbffrak{C}_{l-1}) - \eta\,
       \symbffrak{C}_{l-1}\, ) \notag \\
    &&& \phantom{\symbffrak{C}_{l}} + (\, A^{ \textnormal{C} }
        (\symbffrak{C}_{l}) - \eta\, \symbffrak{C}_{l}\, )
        \,, \label{ACFB.C.l}
  \end{alignat}
\end{subequations}
$\forall l\in \IntegerP$, with $\eta\in ( 0, 2(1-\varpi)\, )$, converges to the solution~\eqref{solution.P.task}, that
is, for any node $n$, $\lim_{l \to \infty} \vect{C}_{l}^{(n)} = \vect{\Phi}_{ \symcal{N} } \vect{\Phi}_{ \symcal{N}
}^{\intercal}$~\cite[Lem.~3.4 and Cor.~3.5]{slavakis2018fejer}. Refer to \cref{thm:C.consensus.rate} for a stronger
result on the linear convergence rate of the sequence of estimates.

\subsection{The proposed distributed value-iteration algorithm}\label{sec:algo}

The aforementioned arguments are consolidated in \cref{algo}.

\begin{algo}[Distributed value iteration (VI)]\label{algo} \mbox{}

  %% \caption{Distributed value iteration (VI)}\label{algo}

  \begin{algorithmic}[1]

    \Require Graph $\symcal{G} = ( \symcal{N}, \symcal{E} )$, data $\{ \symcal{T}^{(n)} \}_{n\in \symcal{N}}$,
    reproducing kernel $\kappa$, $\varpi\in [1/2, 1)$, $\eta\in ( 0, 2(1-\varpi)\, )$, $\gamma \in (\, 0, 1 /
      \norm{\vect{L}}_2 \, ]$, $M\in \IntegerPP$, $J_C \in \{1, \ldots, M\}$.

    \Ensure $(\, \symbffrak{Q}[k] = (\, Q^{(1)} [k], \ldots, Q^{(N)}[k]\, )\, )_{ k\in \IntegerP}$

    % \LineComment{Gather information from neighbors} \LongEq{ \( \vect{C}^{(n)}[0] \coloneqq (1-\lambda) [ \varpi
    % A_{P}^{(n)} ( \boldsymbol{\symcal{P}}[-1] ) + (1 - \varpi) \vect{\Phi}^{(n)} \vect{\Phi}^{(n) \intercal} ] +
    % \lambda \vect{\Phi}^{(n)} \vect{\Phi}^{(n) \intercal} \) \strut}

    \State{Define $\symbffrak{C}_{ \symcal{N} }$ by~\eqref{C.nodes}.}

    \State{Set $\symbffrak{C}_{-1} \coloneqq [ \vect{0}, \ldots, \vect{0}]$. Compute $\symbffrak{C}_0$
      by~\eqref{ACFB.C.zero}.}

    \For{$k = 0, 1, \ldots, \infty$}

      \State{Estimates $\symbffrak{C}_{kM} = (\, \vect{C}_{kM}^{(1)}, \ldots, \vect{C}_{kM}^{(N)}\, )$ are available to
        the agents.}

      \State{Compute $\{ \vect{\Psi}^{(n)}[k] \}_{n=1}^N$ by~\eqref{psi.n.k}.}%
      \label{algo:define.nodal.basis.vectors}

      \State{Compute $\vect{T} (\symbffrak{Q} [k])$ by \eqref{Bmap.in.net}.}

      \State{Set $\symbffrak{Q}_{-1}[k] \coloneqq \symbffrak{Q}[k]$. Compute $\symbffrak{Q}_0 [k]$
        by~\eqref{ACFB.Q.zero}.}

      \For{$m = 0, 1, \ldots, M-1$}\label{algo:consensus.start} \Comment{Run~\eqref{ACFB.Q.m} $M$ times}

        \State{Compute $\symbffrak{Q}_{m+1}[k]$ by~\eqref{ACFB.Q.m}.} \Comment{Info sharing}\label{algo:share.Q}

        \State{Define index $l \coloneqq kM + m$.}\label{algo:def.index.l}

        \If{$(m\! \mod J_C) = 0$}\label{algo:update.P.start}

          \LineComment{Run~\eqref{ACFB.C.l} once every $J_C$ times}

          \State{Compute $\symbffrak{C}_{l+1}$ by~\eqref{ACFB.C.l}.} \Comment{Info sharing}\label{algo:share.cov}

        \Else
        \State{Set $\symbffrak{C}_{l+1} \coloneqq \symbffrak{C}_{l}$ and $\symbffrak{C}_{l} \coloneqq
          \symbffrak{C}_{l-1}$.}
        \EndIf\label{algo:update.P.end}

      \EndFor\label{algo:consensus.end}

      \State{$\symbffrak{Q}[k+1] \coloneqq \symbffrak{Q}_{M}[k]$.}\label{algo:VI.update} \Comment{VI update}

    \EndFor

  \end{algorithmic}
\end{algo}

To establish the connection between~\eqref{ACFB.Q}, \eqref{ACFB.C}, and VI, note that any index $l$ of~\eqref{ACFB.C}
can be expressed as $l = kM + m$ (see line~\ref{algo:def.index.l} of \cref{algo}), where $m \in \{0, 1, \ldots, M-1\}$
is the index of~\eqref{ACFB.Q}, and $k$ is the VI index. This observation emphasizes that~\eqref{ACFB.Q}, which aims to
achieve consensus among Q-functions over $\symcal{G}$, runs only $M$ times between two consecutive VI indices (see
lines~\ref{algo:consensus.start}--\ref{algo:consensus.end} of \cref{algo} and \cref{fig:algo}). Agent $n$ runs
iteration~\eqref{ACFB.C} in parallel with~\eqref{ACFB.Q}. To conserve computational resources and communication
bandwidth, \cref{algo} allows the update in~\eqref{ACFB.C} to be implemented once every $J_C$ iterations ($\IntegerPP
\ni J_C \leq M$); see lines~\ref{algo:update.P.start}--\ref{algo:update.P.end} of \cref{algo}. The effect of $J_C$ on
the performance of \cref{algo} is explored in \cref{fig:pendulum.distance.fixed.point,%
  fig:cartpole.distance.fixed.point}. Iteration~\eqref{ACFB.C} provides~\eqref{ACFB.Q} with estimates $\vect{C}_{l =
  kM}^{(n)}$ through the following update of the nodal basis vectors $\vect{\Psi}^{(n)}[k]$ at VI iteration $k$ (see
also \eqref{psi.n.k.guess}):
\begin{align}
  \vect{\Psi}^{(n)}[k] & \coloneqq (\, \vect{C}_{kM}^{(n)} + \sigma \vect{I}_D\, )^{-1}\, \vect{\Phi}^{(n)}
  \,. \label{psi.n.k}
\end{align}
To justify~\eqref{psi.n.k} recall from the discussion after~\eqref{ACFB.C} that for all sufficiently large values of
$k$, the covariance-matrix estimate $\vect{C}_{kM}^{(n)}$ lies very close to $\vect{\Phi}_{ \symcal{N} } \vect{\Phi}_{
  \symcal{N} }^{\intercal}$. Notice also that the adoption of~\eqref{psi.n.k} in~\eqref{Bmap.nodal} makes $T^{(n)}$
dependent on index $k$. To avoid overloading notations with indices, $k$ will be omitted from $T^{(n)}$ hereafter.

It is clear from the previous discussion that, in addition to the general \cref{ass:setting} and in contrast to most
prior DRL schemes, \cref{algo} also adopts the following assumption.

\begin{assumption}\label{ass:setting.share.P} \textbf{(Covariance-matrix sharing)}
  In \cref{algo}, agent $n$ communicates a copy of its covariance-matrix estimate $\vect{C}_{kM}^{(n)}$ to its neighbors
  $\symcal{N}_n$.
\end{assumption}

Actually, since the $D\times D$ matrix $\vect{C}_{kM}^{( n )}$ is symmetric, only $D (D+1) / 2$ real-valued entries of
$\vect{C}_{kM}^{( n )}$ need to be transmitted to the neighbors $\symcal{N}_n$. Nonetheless, \cref{algo} requires in
principle more communication bandwidth to operate compared to DRL designs that adhere only to
\cref{ass:setting.share.Q}. Surprisingly, the numerical tests in \cref{sec:numerical} reveal the opposite: \cref{algo}
consumes less cumulative communication bandwidth to converge than prior-art DRL designs that follow only
\cref{ass:setting.share.Q}; see \cref{fig:pendulum.episodic,fig:cartpole.episodic}.

\begin{figure}[!t]
  \centering

  \resizebox{\columnwidth}{!}{
    \begin{tikzpicture}
      \usetikzlibrary{decorations.pathreplacing}

      \draw[->, ultra thick, draw=magenta] (0,\ArrayHeight) -- (\ArrayLength,\ArrayHeight) node[below right] {$k$};
      \node[left, text = magenta] at (0,2,0.) {
        \begin{tabular}{c}
          Run\\ VI
        \end{tabular}
      };

      \draw[->, ultra thick, draw=blue] (0,0) -- (\ArrayLength,0) node[below right] {$l$}; \node[left, text = blue] at
      (0,0.) {
        \begin{tabular}{c}
          Run\\ \eqref{ACFB.C}
        \end{tabular}
      };

      \draw[thick] (0, -0.2) -- (0, 0.2);
      \draw[thick] (0, \ArrayHeight+0.2) -- (0,
      \ArrayHeight-0.2);

      \node[below, font = \scriptsize] at (\Scale, 0-0.05) {$M$}; \foreach \x/\k in { \Scale*2/2, \Scale*3/3,
        \ArrayLength-1/K} { \node[below, font = \scriptsize] at (\x, 0-0.05) {$\k M$}; }

      \foreach \x/\k in {\Scale/1, \Scale*2/2, \Scale*3/3, \ArrayLength-1/K} { \draw[thick] (\x, \ArrayHeight-0.05) --
        (\x, \ArrayHeight+0.05); \node[above, font = \scriptsize] at (\x, \ArrayHeight+0.05) {$\k$}; }

      \node[below, font = \scriptsize] at (\Scale*4.5,-0.05) {$l=kM+m$};

      \node at (\Scale*4.5,\PHeight) {$\vect{C}^{(n)}_{kM+m}$};

      \node[below, font = \scriptsize] at (\Scale*4.5+1.5,-0.05) {$\cdots$};

      \node[above] at (3*\Scale+1, \ArrayHeight+0.05) {$\cdots$};

      \node at (3*\Scale+1, \ArrayHeight-0.75) {$\cdots$};

      % \node[below] at (3*\Scale+0.5, -0.05) {$\cdots$};

      \node[above] at (\ArrayLength-0.5, \ArrayHeight+0.05) {$\cdots$};

      % \node at (\ArrayLength-0.5, \ArrayHeight-0.75)
      % {$\cdots$};
      % \node[below] at (\ArrayLength-0.5, -0.05) {$\cdots$};

      % \foreach \x in {\Scale, 2*\Scale, 3*\Scale,
      % \ArrayLength-1} { \draw[->, thick] (\x, 0.2) -- (\x,
      % \ArrayHeight-0.2);
      % }

      \draw[->, thick] (\Scale, 0.2) -- node[right, rotate = 0, font = \scriptsize] {$\vect{C}^{(n)}_{M}$} (\Scale,
      \ArrayHeight-0.2);

      \draw[->, thick] (2*\Scale, 0.2) -- node[right, rotate = 0, font = \scriptsize] {$\vect{C}^{(n)}_{2M}$} (2*\Scale,
      \ArrayHeight-0.2);

      \draw[->, thick] (3*\Scale, 0.2) -- node[right, rotate = 0, font = \scriptsize] {$\vect{C}^{(n)}_{3M}$} (3*\Scale,
      \ArrayHeight-0.2);

      \draw[->, thick] (\ArrayLength-1, 0.2) -- node[right, rotate = 0, font = \scriptsize] {$\vect{C}^{(n)}_{KM}$} (
      \ArrayLength-1, \ArrayHeight-0.2);

      \foreach \x/\k in {\Scale/1, \Scale*2/2, \Scale*3/3, \ArrayLength-1/K} { \draw[decorate, decoration={brace,
            amplitude=5pt}, thick] (\x-\Scale, \SolveHeight + .1) -- (\x, \SolveHeight + .1) node[midway, above=7pt] {};
        \node [above, font = \scriptsize] at (\x-\Scale*0.5, \SolveHeight + .3) {Run \eqref{ACFB.Q}}; }

      \draw[<->, thick] (0,0.9) -- (1.5,0.9);

      \node [above, font = \scriptsize] at (0.75, 0.9) {$M$
        iterations};

    \end{tikzpicture}
  }
  \caption{Iteration~\eqref{ACFB.C} is implemented to provide consensual estimates of the network-wide covariance
    operator~\eqref{Cov.Matrix.global}, while~\eqref{ACFB.Q} provides consensual estimates of the fixed point
    $Q_{\odot}$ of the star-topology B-Map~\eqref{Bmap.star}. Iteration~\eqref{ACFB.C} feeds the covariance-operator
    estimate $\vect{C}_{kM}^{(n)}$ to iteration~\eqref{ACFB.Q} periodically ($l = kM$). This estimate is needed to
    define the nodal basis vectors in~\eqref{psi.n.k} at VI index $k$; see line~\ref{algo:define.nodal.basis.vectors} of
    \cref{algo}. Iteration~\eqref{ACFB.Q} runs only $M$ times between two consecutive VI indices.  } \label{fig:algo}
\end{figure}

\subsection{Performance analysis of \cref{algo}}\label{sec:performance}

First, consider the eigenvalue decomposition (EVD) of the Laplacian matrix $\vect{L} = \vect{U} \diag (\lambda_1,
\ldots, \lambda_{N-1}, \lambda_N ) \vect{U}^{\intercal}$, where $\norm{\vect{L}}_2 = \lambda_1 \geq \ldots \geq
\lambda_{N-1} \geq \lambda_N \geq 0$, and $\vect{U}$ is orthogonal. Because of the connectedness of $\symcal{G}$ by
\cref{ass:setting.topology}, $\lambda_{N-1} > \lambda_N = 0$~\cite[Lem.~4.3]{Bapat.graphs}. The eigenvalue
$\lambda_{N-1}$ is also well known as the algebraic connectivity or Fiedler value of the
graph~\cite{Bapat.graphs}. Define then
\begin{align}
  b_n \coloneqq \frac{\lambda_n}{\lambda_1}\,, \quad \forall n\in \{ 1, \ldots, N\} \,, \label{def.bn}
\end{align}
so that $1 = b_1 \geq b_2 \geq \ldots \geq b_{N-1} > b_{N} = 0$. Define also%
\begin{subequations}%
  \begin{align}
    \varrho (\eta)
    & \coloneqq \max \Bigl \{ \max_{ n \in \symcal{N}
      \setminus \{N\} } \frac{\left\rvert p_n + \sqrt{p_n^2
      + 4q_n} \right\rvert}{2},\, (1 - \eta) \Bigr \}
      \,, \label{varrho.def} \\
    p_n & \coloneqq 2 - \eta - \gamma \lambda_n
          \,, \label{pn.def}\\
    q_n & \coloneqq \varpi
          \gamma \lambda_n + \eta - 1 \,. \label{qn.def}
  \end{align}%
\end{subequations}

\begin{assumptions}\mbox{}
  \begin{asslist}

  \item\label{ass:gamma} Set $\gamma \coloneqq 1 / \norm{\vect{L}}_2 = 1/ \lambda_1$ in \cref{algo}.

  \item\label{ass:contraction} The centralized B-Map $T_{\odot}$ in~\eqref{Bmap.star} is a
    contraction~\cite{hb.plc.book}, that is, Lipschitz continuous with coefficient $\beta_{\odot} \in (0,1)$ and $\norm{
      T_{\odot}(Q_1) - T_{\odot}(Q_2) } \leq \beta_{\odot} \norm{ Q_1 - Q_2 }$, $\forall Q_1, Q_2$. Consequently, it is
    guaranteed that the fixed-point set of $T_{\odot}$ is nonempty and a singleton: $\Fix ( T_{\odot} ) = \{ Q \given
    T_{\odot}(Q) = Q \} = \{ Q_{\odot} \}$~\cite{hb.plc.book}.

  \item\label{ass:bounded.Lipschitz} The Lipschitz coefficients $( \beta^{(n)}[k] )_{k\in \IntegerP}$ of the nodal
    B-Maps $( T^{(n)} = T^{(n)}[k] )_{ k\in \IntegerP }$ in~\eqref{Bmap.nodal} are bounded.

  \item\label{ass:bounded.seq} For any node $n\in \symcal{N}$, sequence $( Q^{(n)}[k] )_{ k\in\IntegerP }$ is bounded.

  \item\label{ass:varpi} Let $\varpi \coloneqq 1/2$ in
    \cref{algo}.

  \item\label{ass:b.N-1} The $b_{N-1}$ of~\eqref{def.bn}
    satisfies $b_{N-1}\in (0, 1/2 )$.

  \end{asslist}
\end{assumptions}

Few comments are in order to justify the introduction of the previous assumptions. By following the arguments in the
proof of Theorem~2 in~\cite{akiyama2024nonparametric}, it can be demonstrated that both $T_{\odot}$ in~\eqref{Bmap.star}
and $T^{(n)} = T^{(n)}[k]$ in~\eqref{Bmap.nodal} are Lipschitz continuous. A detailed discussion on conditions which
ensure that the Lipschitz coefficient of $T_{\odot}$ is strictly smaller than \num{1} (\cref{ass:contraction}) can be
found after Assumptions~3 in~\cite{akiyama2024nonparametric}. To save space, such a discussion and the related proofs
are omitted. It is also worth recalling that the classical Bellman mapping $T_{\diamond}$
in~\eqref{classical.Bellman.inf} is a well-known contraction (in a point-wise sense)~\cite{bertsekas2019reinforcement,
  bertsekas1996neuro}. \cref{ass:bounded.seq} is used to ensure the existence of the constant $C$
in~\eqref{before.induction.Q}. \cref{ass:b.N-1} is taken as a premise to establish
\cref{lemma:b.N-1.max,lemma:h.function}, and to simplify the presentation by avoiding lengthy arguments and proofs in
the general case where $b_{N-1} \in (0, 1]$. Similarly, \cref{ass:gamma} is introduced to simplify proofs.

The following theorem presents the main findings of the performance analysis. \cref{thm:varrho} guarantees that the
ensuing linear convergence rates hold for any topology of the connected graph $\symcal{G}$. \cref{thm:Q.consensus.rate}
asserts that the nodal Q-functions estimates (lines~\ref{algo:consensus.start}--\ref{algo:consensus.end} of \cref{algo})
converge to a consensual Q-function \textit{linearly}~\cite[p.~619]{NocedalWright.book}. A similar result holds true for
the covariance-matrix estimates in \cref{thm:C.consensus.rate}. Moreover, \cref{thm:conv.Q.star} states that for
sufficiently large iteration indices $k$, the difference between the nodal estimate $Q^{(n)}[k]$ and the fixed point
$Q_{\odot}$ of the centralized B-Map $T_{\odot}$---see \cref{sec:central.Bmap}---is bounded by the consensus-step
approximation error. More specifically, this error can be made arbitrarily small at a linear rate with respect to the
parameter $M$ of \cref{algo}. In simple terms, the longer the inner loop
(lines~\ref{algo:consensus.start}--\ref{algo:consensus.end} of \cref{algo}) runs, the smaller the consensus error
becomes, and thus the closer the VI output $Q^{(n)}[k]$ is to $Q_{\odot}$. This ability to render their difference
arbitrarily small indicates that the proposed DRL design closely mirrors the behavior of a centralized node, had such a
node existed---see \cref{algo.central} and \cref{fig:pendulum.distance.fixed.point,fig:cartpole.distance.fixed.point}.

\begin{thm}\label{thm:main} Presume \cref{ass:setting,ass:gamma}. The following hold true.

  \begin{thmlist}

  \item\label{thm:varrho} $\forall \eta\in (0, 2(1 - \varpi)\, )$, $0 < \varrho(\eta) < 1$.

  \item\label{thm:Q.consensus.rate} Let $(\, \symbffrak{Q}_m[k] = [\, Q_m^{(1)}[k], \ldots, Q_m^{(N)}[k]\, ]\, )_{ m\in
    \IntegerP }$ be the sequence generated by~\eqref{ACFB.Q}, $\symbffrak{Q}[k]$ the estimate formed by \cref{algo} at
    the VI index $k$, and $\vect{T}(\cdot)$ defined by~\eqref{Bmap.in.net}. Then, for every node $n\in \symcal{N}$,
    \begin{align*}
      \norm{\, Q_m^{(n)}[k] - \vect{T}( \symbffrak{Q}[k] )\, \vect{1}_N \, } = \symcal{O}(\, m\, \varrho^m(\eta)\, ) \,,
    \end{align*}
    where $\norm{\cdot}$ stands for the Euclidean norm in $\Real^D$, and $\symcal{O}(\cdot)$ is the classical big-oh
    notation~\cite{Apostol}.

  \item\label{thm:C.consensus.rate} Let $( \symbffrak{C}_l = [\, \vect{C}_l^{(1)}, \ldots, \vect{C}_l^{(N)}\, ] )$ be
    the sequence generated by~\eqref{ACFB.C}, and $\vect{\Phi}_{ \symcal{N} } \vect{\Phi}_{ \symcal{N} }^{\intercal}$
    the network-wide covariance matrix of~\eqref{Cov.Matrix.global}. Then, for every node $n\in \symcal{N}$,
    \begin{align*}
      \norm{\, \vect{C}_l^{(n)} -  \vect{\Phi}_{ \symcal{N} } \vect{\Phi}_{ \symcal{N} }^{\intercal} \, }_{
        \textnormal{F} } = \symcal{O}(\, l\, \varrho^l(\eta)\, ) \,.
    \end{align*}

  \item\label{thm:conv.Q.star} Consider also \cref{ass:contraction,ass:bounded.seq}. Then, there exists $C\in \RealPP$
    such that (s.t.)
    \begin{align*}
      \lim\sup_{k\to \infty} \norm{\, Q^{(n)} [k] - Q_{\odot} \, } \leq C\, \frac{1}{1 - \beta_{\odot}}\, M\, \varrho^M
      (\eta) \,,
    \end{align*}
    where $Q_{\odot}$ is the unique fixed point of the centralized B-Map $T_{\odot}$---see \cref{ass:contraction} and
    \cref{algo.central}.
  \end{thmlist}

\end{thm}

\begin{IEEEproof}
  See the appendix.
\end{IEEEproof}

Interestingly, the following theorem states that the optimal learning rate $\eta_*$ for recursions \eqref{ACFB.Q} and
\eqref{ACFB.C}, which offers the ``fastest'' linear convergence in \cref{thm:main}, is determined by the value $b_{N-1}$
in~\eqref{def.bn}. Although the statements of \cref{thm:main} hold true for any topology of the connected graph
$\symcal{G}$, the following ``optimal'' learning rate depends on the graph topology through the quantity $b_{N-1}$.

\begin{thm}\label{thm:optimal.eta}
  Consider \cref{ass:gamma,ass:varpi,ass:b.N-1}. Notice that under \cref{ass:varpi}, $\eta \in (0, 1)$. Moreover, recall
  from \eqref{def.bn} that $b_{N-1} \coloneqq \lambda_{N-1} / \lambda_1$, with $\lambda_{N-1} \in \RealPP$ being the
  algebraic connectivity or Fiedler value of the graph. The optimal learning rate $\eta_*$ for \eqref{ACFB.Q} and
  \eqref{ACFB.C} becomes
  \begin{align*}
    \eta_* & \coloneqq \arg \min_{ \eta \in (0, 1) } \varrho( \eta ) = - b_{N-1} + \sqrt{ 2 b_{N-1} } \,.
  \end{align*}
\end{thm}

\begin{IEEEproof}
  See the appendix.
\end{IEEEproof}

\section{Numerical Tests}\label{sec:numerical}

To validate \cref{algo}, a network $\symcal{G}$ with $N = 25$ nodes arranged on a $5 \times 5$ orthogonal grid is
used. Each agent is placed at a node $n \in \{ 1, \dots, 25 \}$ of $\symcal{G}$, where agents communicate with their
neighbors to the north, south, east, and west. Each of the \num{25} agents is assigned an independent system and
learning task, resulting in a total of \num{25} systems. Two scenarios are considered: one where each system is a
pendulum~\cite{openai, pendulum.software, Shil:DELCON:22} (\cref{sec:pendulum}) and another where each system is a
cartpole~\cite{openai, cartpole.software} (\cref{sec:cartpole}); see \cref{fig:envs}. The goal is for all agents to
collaborate via the graph topology to efficiently complete their learning tasks with minimal communication cost. Both
considered scenarios involve discrete and even categorical action spaces; extending the proposed framework to continuous
action spaces is left for future work. Although the proposed framework can accommodate any graph topology (see
\cref{ass:setting.topology}), tests for a star-topology graph are deferred to future work, as they overlap with the
domain of federated RL~\cite{Zhang:2024:FedSARSA, Lan:2025:AFedPG}. A comprehensive comparison of \cref{algo.central}
with the broad topic of federated RL lies beyond the scope of the present manuscript and therefore warrants a dedicated
publication.

\cref{algo} competes against the following designs.

\begin{enumerate}[label=\textbf{(\roman*)}]

\item \textbf{(D-FQ)} The decentralized fitted Q-iteration (D-FQ)~\cite{distributed.TD} solves the TD
  task~\eqref{Bellman.TD}. In its original form, \cite{distributed.TD} assumes that all agents share the same state
  information, \ie, $\vect{s}_i^{(n)} = \vect{s}_i^{(n^{\prime})}$, for all $i$ and for all $n, n^{\prime} \in
  \symcal{N}$ (global state space). However, since \cref{ass:setting.trajectory} relaxes this constraint, allowing
  agents to keep their states private, \cite{distributed.TD} is adapted to the current setting by eliminating the
  assumption of a global state space.

\item \textbf{(D-LSTD)} The diffusion off-policy gradient TD~\cite{distributed.LSTD} efficiently minimizes, via
  stochastic gradient descent, a primal-dual reformulation of the widely used projected Bellman residual error (PBRE)
  encountered in the classical LSTD~\cite{bertsekas2019reinforcement}. Due to the use of PBRE, the acronym D-LSTD will
  be used hereafter to refer to~\cite{distributed.LSTD}. Originally, \cite{distributed.LSTD} was designed for J- and not
  Q-functions, and for online/streaming data. However, in the current setting, as described by \cref{ass:setting}, where
  the data is fixed, each gradient-TD step of D-LSTD is performed using all the available data at agent $n$ (batch
  processing), for a total of $M$ steps, similar to \cref{algo}. Moreover, to robustify the original policy improvement
  of~\cite{distributed.LSTD}, the following running-average ``smoothing'' strategy is employed: $\mu^{(n)}[k+1] (
  \vect{s} )= \argmin_{ a \in \symcal{A} }Q^{(n)}_{ \textnormal{smooth} }[k] (\, \vect{z}( \vect{s}, a)\, )$, $\forall
  \vect{s}$, where $Q^{(n)}_{\textnormal{smooth}}[k] = 0.3\, Q^{(n)}_{\textnormal{smooth}} [k-1] + 0.7\, Q^{(n)}[k]$.

\item \textbf{(Gossip-NN)} The Gossip-based~\cite{NN.gossip}, originally designed for general distributed learning
  tasks, can also be applied to the TD task~\eqref{Bellman.TD}, where a fully connected neural network (NN) serves as
  the nonlinear Q-function. Consequently, \cite{NN.gossip} is referred as Gossip-NN hereafter. Notably, the use of an NN
  makes Gossip-NN \textit{parametric,} distinguishing it from the proposed nonparametric \cref{algo}.

\item \textbf{(D-TD[ADMM])} An ADMM-based~\cite{boyd2011distributed} solution to the TD task \eqref{Bellman.TD},
  proposed for the first time here to compete against the iterations \eqref{ACFB.Q} and \eqref{ACFB.C}. Henceforth,
  D-TD[ADMM] will be used to denote this ADMM-based solution. The regularization term $\sigma^{\prime} \norm{ Q^{(n)}
  }_{ \symcal{H} }^2$, $\sigma^{\prime}\in \RealPP$, is also included in the loss $\symcal{L}_{ \text{TD} }^{(n)}$ of
  \eqref{Bellman.TD} to mimic the regularization offered by $\sigma$ in \eqref{psi.n.k}, and to stabilize the
  iterations. D-TD[ADMM] employs also RFFs---see \eqref{RFF}---for dimensionality reduction.

\end{enumerate}

The parameters for each method were carefully tuned, and the curves corresponding to the parameters that produced the
``best'' performance for each method are shown in the following figures. Each curve represents the uniformly averaged
result of \num{100} independent tests.

In \cref{algo}, $\sigma = 0.01$ in~\eqref{psi.n.k}, $M=50$, while $\varpi = 0.5, \gamma = 1 / \norm{ \vect{L} }_2$ for
the parameters of~\cite{slavakis2018fejer} in \cref{algo}, and $\eta = -b_{N-1} + (2b_{N-1})^{1/2}$ according to
\cref{thm:optimal.eta}. The discount factor $\alpha = 0.9$ is used for all employed methods. Moreover, for all
competitors of \cref{algo}, dimension $D$ of the RFF approximating space is set as $D = 500$ for \cref{sec:pendulum},
while $D = 250$ for \cref{sec:cartpole}. Several values of $D$ will be explored for \cref{algo}.

All employed methods run their distributed algorithm for $M$ iterations between two consecutive value-iteration steps,
as shown in \cref{fig:algo}. The value of $M$, determined through extensive tuning, varies between methods and is listed
in \cref{tab:M}. Note that \cref{algo} uses the smallest value of $M$, as it requires fewer iterations than its
competitors to reach consensus, as demonstrated in \cref{fig:pendulum.consensus} and \cref{fig:cartpole.consensus}, and
further supported by \cref{thm:Q.consensus.rate}.

Because each competing method employs a different value of $M$, selected after extensive fine-tuning for reaching
optimal performance, adopts distinct algorithmic strategies for sharing varying amounts of information among agents, and
seeks to minimize communication costs while meeting its objectives, the curves in
\cref{fig:pendulum.episodic,fig:pendulum.distance.fixed.point,fig:cartpole.episodic,fig:cartpole.distance.fixed.point}
are presented in a nonstandard manner to ensure fairness in comparisons. Rather than plotting the loss functions against
the VI iteration indices, each point on the curves represents the loss as a function of the cumulative communication
cost (in bytes) incurred across $\symcal{G}$. For clarity, lines~\ref{algo:share.Q} and~\ref{algo:share.cov} of
\cref{algo} indicate the precise locations where information exchange, and thus communication cost, occurs in the
proposed framework. As a general guideline, curves positioned closer to the left and bottom edges of those figures
correspond to superior performance.

\begin{table}[ht!]
  \begin{center}
    \caption{Values of $M$ per method and
      scenario} \label{tab:M}
    \begin{tabular}{|l|r|r|} \hline
      Method $\setminus$ Scenario & Pendulum & Cartpole
      \\ \hline
      D-FQ~\cite{distributed.TD} & \num{500} & \num{500} \\
      D-LSTD~\cite{distributed.LSTD} & \num{2500} & \num{2500}
      \\
      Gossip-NN~\cite{NN.gossip} & \num{1000} & \num{2000}
      \\
      D-TD[ADMM] & \num{2000} & \num{2000} \\
      \cref{algo} & \num{50} & \num{50} \\
      \hline
    \end{tabular}
  \end{center}
\end{table}

\subsection{Network of pendulums}\label{sec:pendulum}

\begin{figure}[t]

  \subfloat[Pendulum]{\centering\includegraphics[width = .45\columnwidth ]{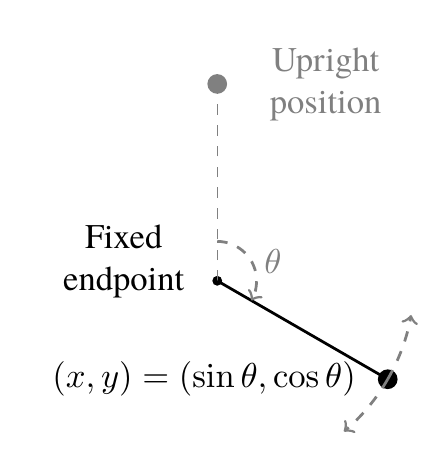} \label{fig:pendulum} }%
  \subfloat[Cartpole]{\centering\includegraphics[width = .45\columnwidth ]{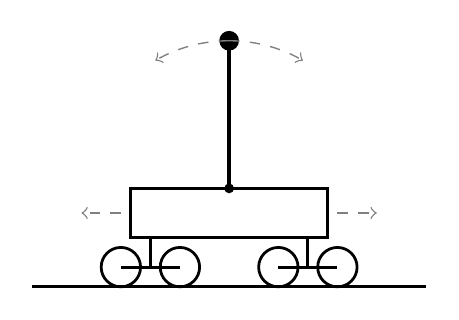} \label{fig:cartpole}
  }

  \caption{Software for the pendulum and cartpole environments can be found in~\cite{pendulum.software}
    and~\cite{cartpole.software}, respectively.} \label{fig:envs}
\end{figure}

Each of the \num{25} agents is assigned a pendulum~\cite{openai, pendulum.software, Shil:DELCON:22}, for a total of
\num{25} pendulums. One endpoint of each pendulum is fixed, while the other is free to move, as illustrated in
\cref{fig:pendulum}. At node $n$ of $\symcal{G}$, agent $n$ applies torque to pendulum $n$, and by sharing information
with neighboring agents, the goal is for all pendulums to collectively swing from their bottom (rest) position to the
upright position and remain there, with the minimal possible communication cost.

According to~\cite{openai}, the generic state at node $n$ is defined as $\vect{s}^{(n)} \coloneqq [ \sin \theta^{(n)},
  \cos \theta^{(n)}, \dot{\theta}^{(n)} ]^{\intercal} \in \symcal{S} \coloneqq [-1, 1] \times [-1, 1] \times \Real$,
where $\theta^{(n)}$ measures the angle between the current direction of the pendulum's arm and the upward direction,
$\dot{\theta}^{(n)}$ is the angular velocity, while torque serves as action $a^{(n)} \in \symcal{A}$, with the action
space $\symcal{A}$ defined as the finite grid resulting from evenly dividing $[-2, 2]$ into \num{10} equal intervals.

The mapping $\vect{z}(\cdot, \cdot)$ of \cref{sec:preliminaries} takes here the following simple form: $\vect{z}(\cdot,
\cdot) \colon \symcal{S} \times \symcal{A} \to \Real^4 \colon (\vect{s}^{(n)}, a^{(n)}) \mapsto \vect{z}(
\vect{s}^{(n)}, a^{(n)} ) \coloneqq [ \vect{s}^{(n) \intercal}, a^{(n)} ]^{\intercal} \eqqcolon
\vect{z}^{(n)}$. Moreover, the one-step loss function $g(\cdot)$, or, equivalently, the one-step reward $-g(\cdot)$, is
defined as
\begin{align*}
  g( \vect{z}^{(n)} ) \coloneqq ( \theta^{(n)} )^2 + 0.1\, ( \dot{\theta}^{(n)} )^2 + 0.001\, ( a^{(n)} )^2 \,.
\end{align*}
Notice that any deviation from the upright position, $\theta^{(n)} \neq 0$, in conjunction with nonzero angular velocity
$\dot{\theta}^{(n)}$ and applied torque $a^{(n)}$, is strongly penalized by the quadratic law of $g( \cdot
)$. Accordingly, the agent is incentivized to select actions that minimize this penalization.

The data trajectory $\symcal{T}^{(n)}$ of \cref{ass:setting.trajectory} is generated inductively as follows: starting
with a random $\vect{s}_0^{(n)}$ as in~\cite{openai}, at state $\vect{s}_i^{(n)}$, action $a_i^{(n)}$ is selected
randomly from $\symcal{A}$, and receives the one-step loss $g_i^{(n)} = g( \vect{z}_i^{(n)} )$ to transition to
$\vect{s}_{i+1}^{(n)} \coloneqq \vect{s}_i^{(n)\prime}$, according to a transition module function
$F_{\textnormal{trans}} (\cdot)$, inherent to the system~\cite{openai, pendulum.software}. Although
$F_{\textnormal{trans}} (\cdot)$ does not include any noise in its original design~\cite{openai}, to offer a more
realistic setting here, measurement noise is also considered, so that $( \theta_{i+1}^{(n)}, \dot{\theta}_{i+1}^{(n)} )
= F_{\textnormal{trans}} (\theta_i^{(n)} + \epsilon_1 , \dot{\theta}_i^{(n)} + \epsilon_2, a_i^{(n)} + \epsilon_3 )$,
where $\epsilon_k$ is a random variable that follows the Gaussian PDF $\symcal{N} (0, \sigma_k)$, with $\sigma_1 = 0.05,
\sigma_2 = 0.25$, and $\sigma_3 = 0.05$. This inductive construction of the trajectory continues till index $i$ reaches
the number $N_{ \textnormal{av} }^{(n)} = 500$.

To validate the current estimate $Q^{(n)}[k]$ of each one of the employed methods, \textit{test}\/ or
\textit{episodic}\/ trajectory data $\symfrak{E}_k \coloneqq ( \symfrak{s}_i^{(n)}[k], \symfrak{a}_i^{(n)}[k],
\symfrak{s}_{i+1}^{(n)}[k] )_{i=0}^{ N_{\text{e}} - 1 }$, for some $N_{\text{e}}\in \IntegerPP$, are generated
inductively as follows: starting from the pendulum's rest position $\symfrak{s}_{0}^{(n)}[k] \coloneqq [ \sin \pi,
  \cos\pi, 0 ]^{\intercal}$, and given $\symfrak{s}_i^{(n)}[k]$, apply torque $\symfrak{a}_i^{(n)}[k] \coloneqq \arg
\min_{ a \in \symcal{A} } Q^{(n)}[k] ( \symfrak{s}_i^{(n)}[k], a )$ according to~\eqref{def.action} for the pendulum to
swing to its new state $\symfrak{s}_{i+1}^{(n)}[k]$ via the earlier met transition module function
$F_{\textnormal{trans}} (\cdot)$. Noise is \textit{not}\/ considered in the implementation of $F_{\textnormal{trans}}
(\cdot)$, unlike the case of training data generation. The reason is that the current estimate $Q^{(n)}[k]$, despite the
fact that it was learned from noisy training data, needs to be validated on noiseless, actual, or ground-truth
data. Eventually, the quality of $Q^{(n)}[k]$ is validated by the following ``episodic loss''
\begin{align}
  \symcal{L}_{ \textnormal{e} } [k] \coloneqq \tfrac{1}{N N_{\text{e}}} \sum_{n\in\symcal{N}} \sum_{i=0}^{ N_{\text{e}}
    - 1 } g( \symfrak{s}_i^{(n)}[k], \symfrak{a}_i^{(n)}[k] ) \,. \label{episodic.loss}
\end{align}
For the current scenario, $N_{\text{e}} = 200$.

\begin{figure}[t!]
  \def\RATIO{.8}
  \def\MkSIZE{3}
  \def\BsLINE{-.5}
  \centering

  \subfloat[ Episodic loss~\eqref{episodic.loss} ]{%
    \includegraphics[width = \RATIO\columnwidth]{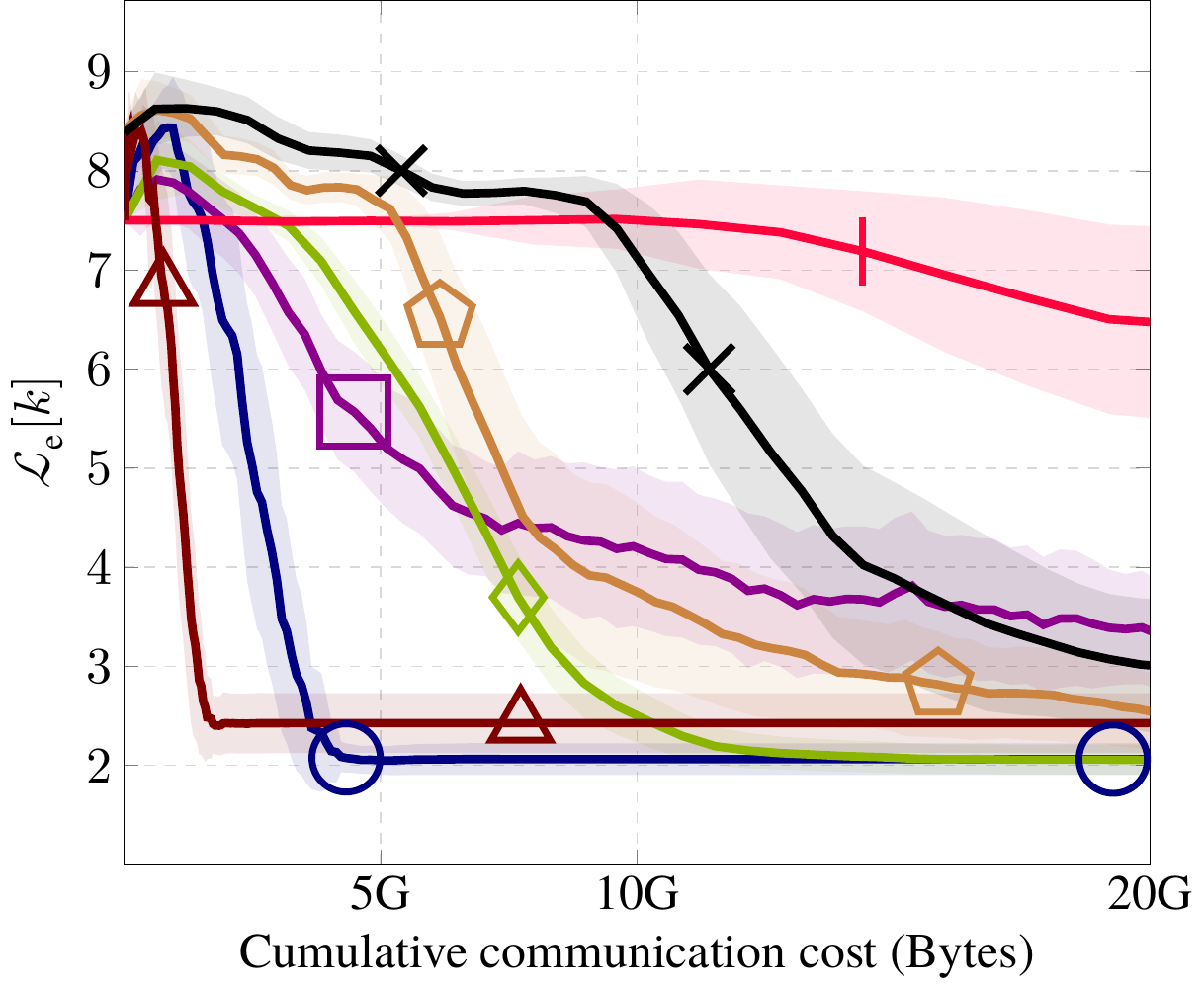}\label{fig:pendulum.episodic}%
  }\\
  \subfloat[ Distance~\eqref{distance.to.fp} for different values of $J_C$ in \cref{algo} ]{%
    \includegraphics[width = \RATIO\columnwidth]{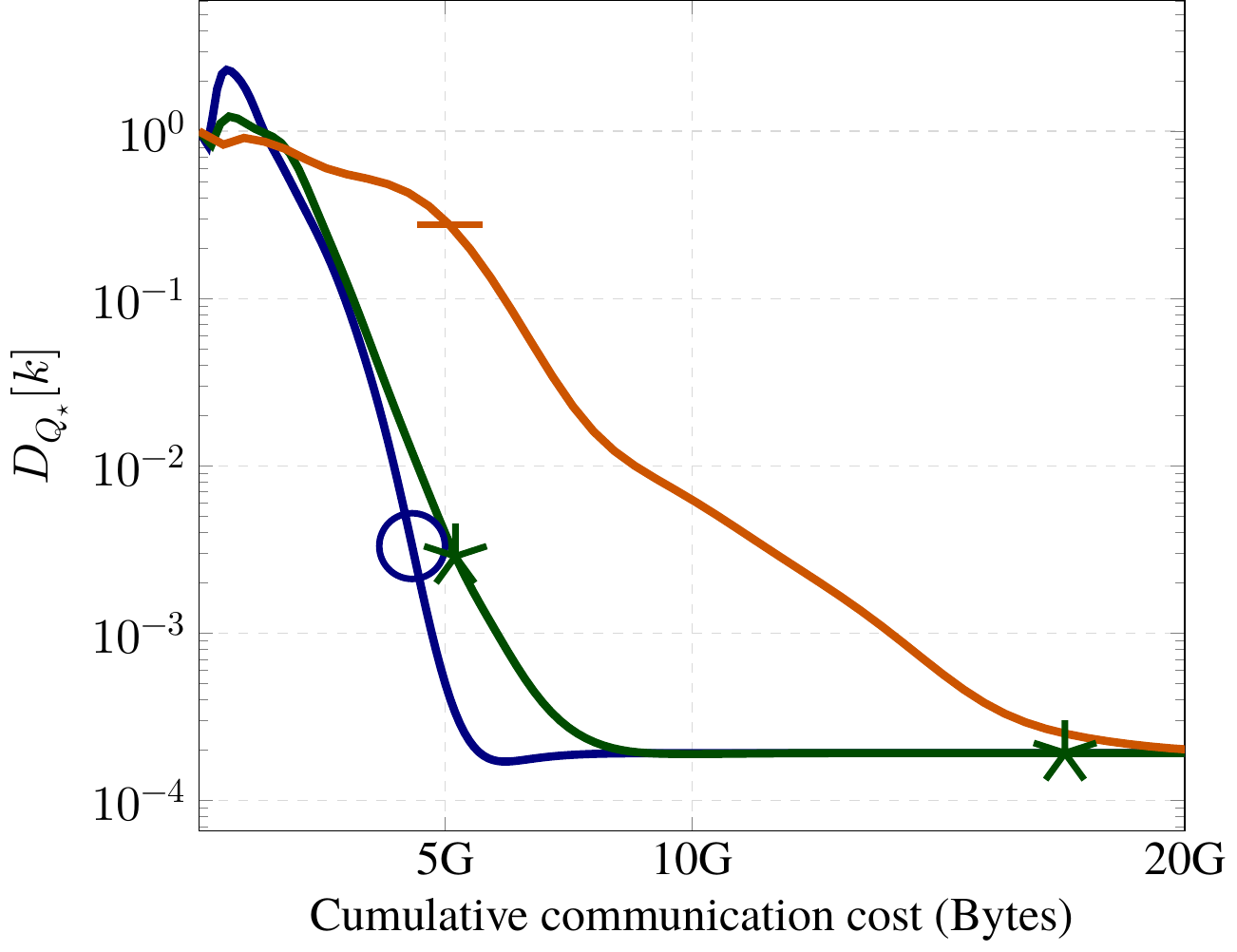}\label{fig:pendulum.distance.fixed.point}%
  }\\
  \subfloat[ Consensus losses~\eqref{consensus.loss} and \eqref{consensus.loss.nn} vs.\ $kM+m$, where the running
    indices are $(k, m)$: the outer VI index $k = 0, 1, \ldots$, and the inner consensus index $m = 0, \ldots, M-1$ ]{
    \includegraphics[width = \RATIO\columnwidth]{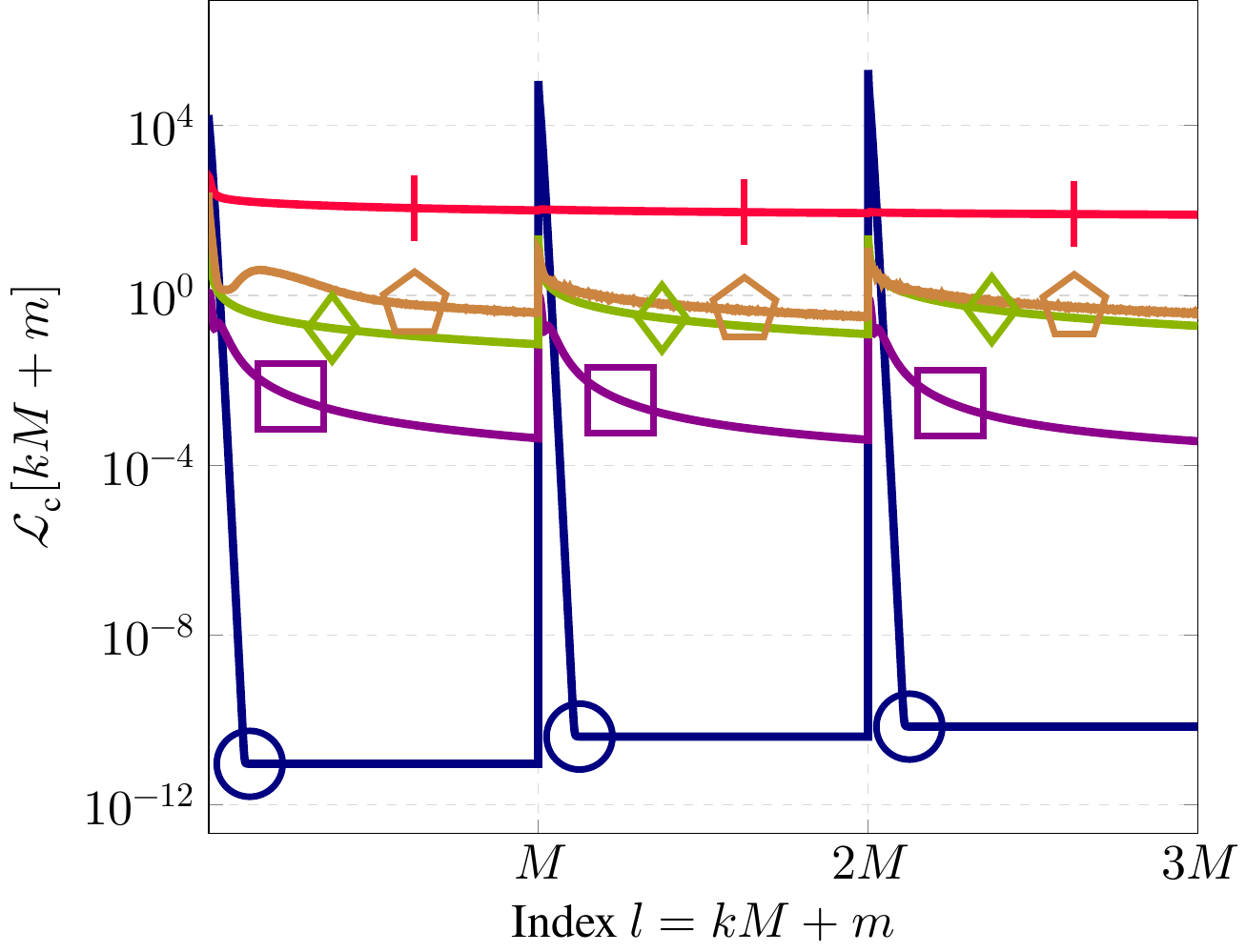}\label{fig:pendulum.consensus} }

  \caption{Network of pendulums (\cref{sec:pendulum}). D-FQ~\cite{distributed.TD}:~\protect \tikz[baseline = \BsLINE ex]
    { \protect \node[mark size = \MkSIZE pt, color = darkmagenta ] {\protect \pgfuseplotmark{square}}; }\,,
    D-LSTD~\cite{distributed.LSTD}:~\protect \tikz[baseline = \BsLINE ex]{ \protect \node[mark size = \MkSIZE pt, color
        = americanrose ] {\protect \pgfuseplotmark{|}}}, Gossip-NN~\cite{NN.gossip} (NN with \num{506}
    parameters):~\protect \tikz[baseline = \BsLINE ex]{ \protect \node[mark size = \MkSIZE pt, color = peru ] {\protect
        \pgfuseplotmark{pentagon}}}\,, Gossip-NN~\cite{NN.gossip} (NN with \num{938} parameters):~\protect
    \tikz[baseline = \BsLINE ex]{ \protect \node[mark size = \MkSIZE pt, color = black ] {\protect
        \pgfuseplotmark{x}}}\,, D-TD[ADMM]:~\protect \tikz[baseline = \BsLINE ex]{ \protect \node[mark size = \MkSIZE
        pt, color = applegreen ] {\protect \pgfuseplotmark{diamond}}}\,, \cref{algo} $(D, J_C) = (500, 50)$:~\protect
    \tikz[baseline = \BsLINE ex]{ \protect \node[mark size = \MkSIZE pt, color = navy ] {\protect
        \pgfuseplotmark{o}}}\,, \cref{algo} $(D, J_C) = (500, 25)$:~\protect \tikz[baseline = \BsLINE ex]{ \protect
      \node[mark size = \MkSIZE pt, color = darkgreen ] {\protect \pgfuseplotmark{star}}}\,, \cref{algo} $(D, J_C) =
    (500, 10)$:~\protect \tikz[baseline = \BsLINE ex]{ \protect \node[mark size = \MkSIZE pt, color = burntorange ]
      {\protect \pgfuseplotmark{-}}}\,, \cref{algo} $(D, J_C) = (300, 50)$:~\protect \tikz[baseline = \BsLINE ex]{
      \protect \node[mark size = \MkSIZE pt, color = maroon ] {\protect \pgfuseplotmark{triangle}}}\,. The shaded areas
    in \cref{fig:pendulum.episodic} correspond to values in the range of $(\text{mean}) \pm 0.5 \times (\text{standard
      deviation})$. The $M$-periodic jumps observed in \cref{fig:pendulum.consensus} occur because the algorithms
    perform $M$ consensus steps (index $m$) across the graph before each VI update (index
    $k$).} \label{fig:pendulum.losses}
\end{figure}

\cref{algo} is also validated via the normalized distance
\begin{align}
  D_{ Q_{\odot} } [k] \coloneqq \tfrac{1}{N} \sum_{ n \in \symcal{N} } \frac{ \norm{ Q^{(n)}[k] - Q_{\odot} }^2 } {
    \norm{Q_{\odot} }^2 } \label{distance.to.fp}
\end{align}
to a fixed point $Q_{\odot}$ of the star-topology map $T_{\odot}$ defined in~\eqref{Bmap.star}. However, in general,
$Q_{\odot}$ cannot be obtained in closed form from~\eqref{Bmap.star}. Assuming that $T_{\odot}$ is a contraction
mapping, $Q_{\odot}$ is taken to be the limit point of the Banach-Picard iteration~\cite{hb.plc.book}: for an
arbitrarily fixed $Q_0$, $Q_{k+1} \coloneqq T_{\odot} (Q_k)$, $\forall k\in \IntegerP$.

To assess whether consensus is achieved by the employed algorithms, the following ``consensus loss'' is considered:
\begin{align}
  \symcal{L}_{ \text{c} } [kM + m] \coloneqq \tfrac{1}{N(N-1)}
  \sum_{ n \neq n^{\prime} }
  \norm{\, Q_m^{(n)}[k] - Q_m^{ (n^{\prime}) }[k]\,
  }  \,, \label{consensus.loss}
\end{align}
where $k\in \IntegerP$ and $m\in \{0, \ldots, M-1\}$. However, for Gossip-NN~\cite{NN.gossip}, where
dense NNs are used, the following consensus loss is adopted:
\begin{alignat}{2}
  & \symcal{L}_{\text{c}}^{\text{NN}} [ kM+m ]
  && \notag \\
  & \coloneqq \tfrac{1}{ N(N-1) } \sum_{ n \neq n^{\prime}
    } \Bigl ( && \sum\nolimits_{i=1}^{ L_\text{NN} }
                 \norm{ \vect{W}_i^{(n)} - \vect{W}_{i}^{
                 (n^{\prime}) } }_{\text{F}}^2 \notag \\
  &&& + \norm{ \vect{b}_i^{(n)} - \vect{b}_i^{ (n^{\prime})
      } }^2 \Bigr )^{1/2} \,, \label{consensus.loss.nn}
\end{alignat}
where $\vect{W}_i^{(n)}$ and $\vect{b}_i^{(n)}$ stand for the matrix of weights and vector of offsets of the $i$th NN
layer at node $n$, respectively.

\cref{fig:pendulum.episodic} shows that \cref{algo}, with $(D, J_C) = (500, 50)$ and $(D, J_C) = (300, 50)$, outperforms
all other methods in terms of the episodic loss~\eqref{episodic.loss}, as these configurations are positioned closest to
the left and bottom edges of the figure. However, a trade-off arises. Reducing the RFF dimension $D$ from \num{500} to
\num{300} decreases the cumulative communication cost needed for the curve to reach its ``steady state,'' since fewer
parameters are communicated among agents. On the other hand, the value of the steady-state loss is increased, as using
fewer parameters reduces the RFF space's ability to adequately approximate Q-functions.

It is also important to note that D-TD[ADMM], introduced here to solve \eqref{Bellman.TD}, achieves the same loss-value
level as \cref{algo} with $(D, J_C) = (500, 50)$, but at the cost of significantly higher communication (more than
double). Recall that in D-TD[ADMM], agents only communicate their Q-function information. Additionally, increasing the
number of NN parameters in Gossip-NN ``delays'' convergence to a steady state, as more parameters are communicated among
agents over $\symcal{G}$ per VI iteration. However, this increase leads to a slight improvement in the steady-state
loss value, due to the enhanced Q-function approximation capacity provided by the larger number of NN parameters.

\cref{fig:pendulum.distance.fixed.point} illustrates the effect of the parameter $J_C$ in \cref{algo} on the distance
loss~\eqref{distance.to.fp}. The curves confirm that as $J_C$ increases, $\vect{C}_l^{(n)}$ is shared less frequently
among neighbors via \eqref{Ap.n} in the computation of \eqref{ACFB.C.l}, resulting in a smaller communication cost
footprint. However, the robustness of \cref{algo} to changes in $J_C$ is noteworthy: the steady-state loss value appears
unaffected by these variations.

Although each method employs different values of $M$ to achieve consensus among agents between VI iterations (see
\cref{fig:algo} and \cref{tab:M}), to assess the consensus quality on a common platform, $M$ is set to \num{2000} in
\cref{fig:pendulum.consensus}. It is evident that \cref{algo} achieves consensus quickly with low loss values in
\eqref{consensus.loss}, supported theoretically by \cref{thm:Q.consensus.rate}. This justifies the choice of $M = 50$ in
\cref{tab:M}, as there is no need to wait for \num{2000} iterations before progressing to the next VI recursion (see
\cref{fig:algo}).

\subsection{Network of cartpoles}\label{sec:cartpole}

Similar to the setup in \cref{sec:pendulum}, a cartpole~\cite{openai, cartpole.software} is assigned to each of the
\num{25} agents on the $5 \times 5$ grid. A cartpole consists of a cart and a pole (\cref{fig:cartpole}), with one end
of the pole attached to the cart, which moves horizontally on a straight line, while the other end is free to
move. Following \cite{openai, cartpole.software}, the state of the cartpole at node $n$ is represented by the tuple
$(x^{(n)}, v^{(n)}, \theta^{(n)}, \dot{\theta}^{(n)}) \in \symcal{S} \coloneqq \Real \times \Real \times [-\pi, \pi]
\times \Real$, where $x^{(n)}$ is the horizontal position of the cart, $v^{(n)}$ is the cart’s velocity, $\theta^{(n)}$
is the angle between the pole’s current direction and its upright position (similar to the pendulum case), and
$\dot{\theta}^{(n)}$ is the angular velocity. Actions here are not numerical but categorical: either move the cart to
the left, $a = \text{L}$, or move the cart to the right, $a = \text{R}$, by applying some pre-defined force. In other
words, $\symcal{A} \coloneqq \{ \text{L}, \text{R}\}$. The objective of each individual agent is to select actions that
move the cart horizontally so that $-B_x \leq x^{(n)} \leq B_x$ and $-B_{\theta} \leq \theta^{(n)} \leq B_{\theta}$, for
some $B_x, B_{\theta} \in \RealPP$~\cite{openai, cartpole.software}. The collective objective is for all agents to
collaborate by exchanging information with their neighbors to achieve their individual goals with the least possible
communication cost.

Following the strategy of~\cite{xu2007klspi} to amplify separability between actions, the mapping $\vect{z}(\cdot,
\cdot)$ of \cref{sec:preliminaries} takes here the following form: $\vect{z}( \cdot, \cdot) \colon \symcal{S} \times
\symcal{A} \to \Real^8 \colon (\vect{s}^{(n)}, a^{(n)}) \mapsto \vect{z}( \vect{s}^{(n)}, a^{(n)} ) \eqqcolon
\vect{z}^{(n)}$ with 
\begin{align}
  \vect{z}^{(n)} \coloneqq
  \begin{cases}
    [ \frac{x^{(n)}}{4}, \frac{v^{(n)}}{4}, \theta^{(n)}, \frac{ \dot{\theta}^{(n)} }{4}, 0, 0, 0, 0 ]^{\intercal} \,, &
    \text{if}\ a = \text{L} \,, \\
    [ 0, 0, 0, 0, \frac{x^{(n)}}{4}, \frac{v^{(n)}}{4}, \theta^{(n)}, \frac{\dot{\theta}^{(n)}}{4} ]^{\intercal}\,, &
    \text{if}\ a = \text{R} \,, \label{z.cartpole}
  \end{cases}
\end{align}
where scaling by $1/4$ was introduced to facilitate learning. Moreover, the one-step loss $g(\cdot)$, or, equivalently,
the one-step reward $-g(\cdot)$, is defined by
\begin{align*}
  g( \vect{z}^{(n)} )
  \coloneqq \begin{cases}
    0\,, & \text{if}\ \lvert x^{(n)} \rvert > B_x\ \text{or}\ \lvert \theta^{(n)} \rvert > B_{\theta}\,, \\
    -1\,, & \text{otherwise}\,.
  \end{cases}
\end{align*}

\begin{figure}[t!]
  \def\RATIO{.8}
  \def\MkSIZE{3}
  \def\BsLINE{-.5}
  \centering

  \subfloat[ Episodic loss~\eqref{episodic.loss} ]{%
    \includegraphics[width = \RATIO\columnwidth]{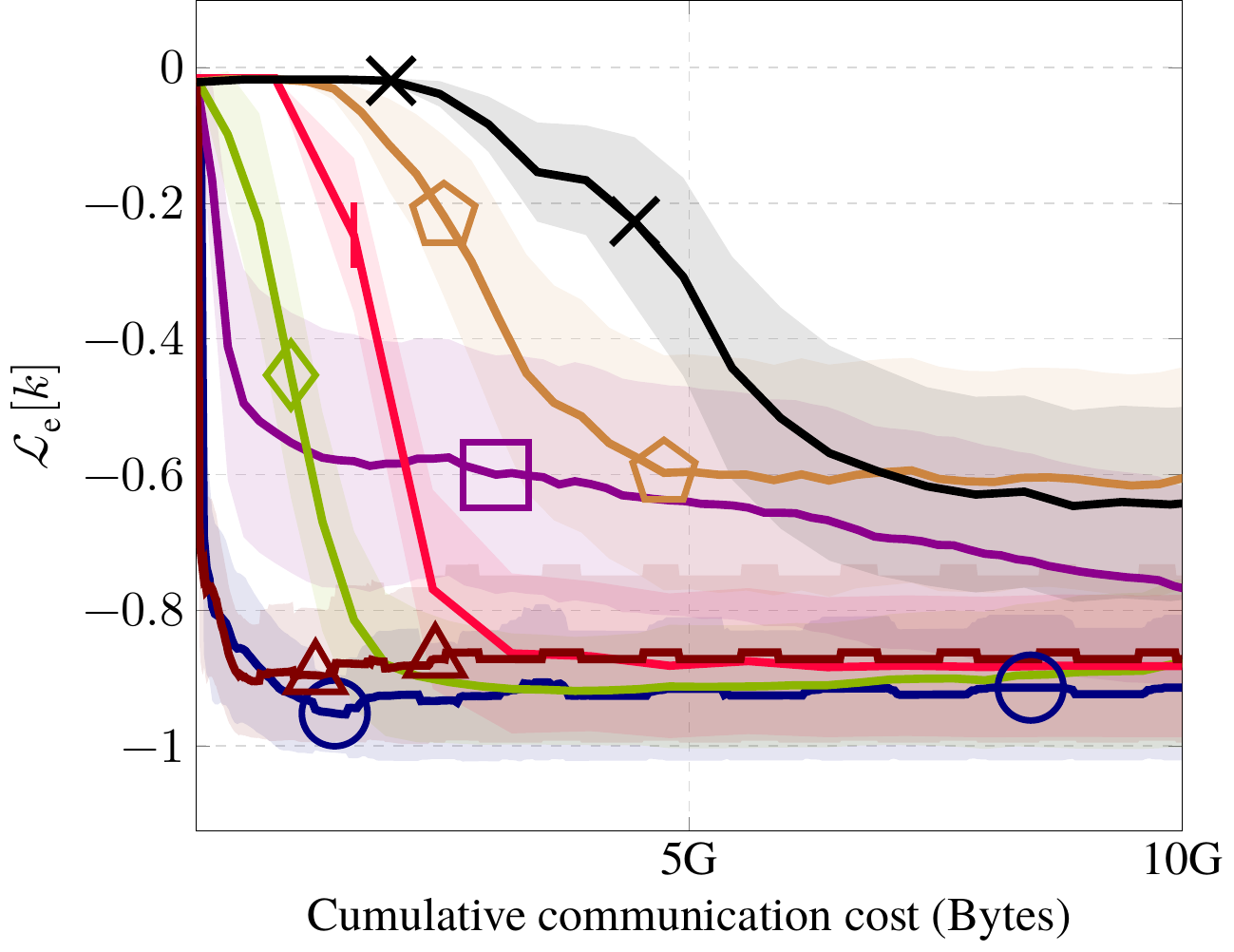}\label{fig:cartpole.episodic}%
  }\\
  \subfloat[ Distance~\eqref{distance.to.fp} for different values of $J_C$ in \cref{algo} ]{%
    \includegraphics[width = \RATIO\columnwidth]{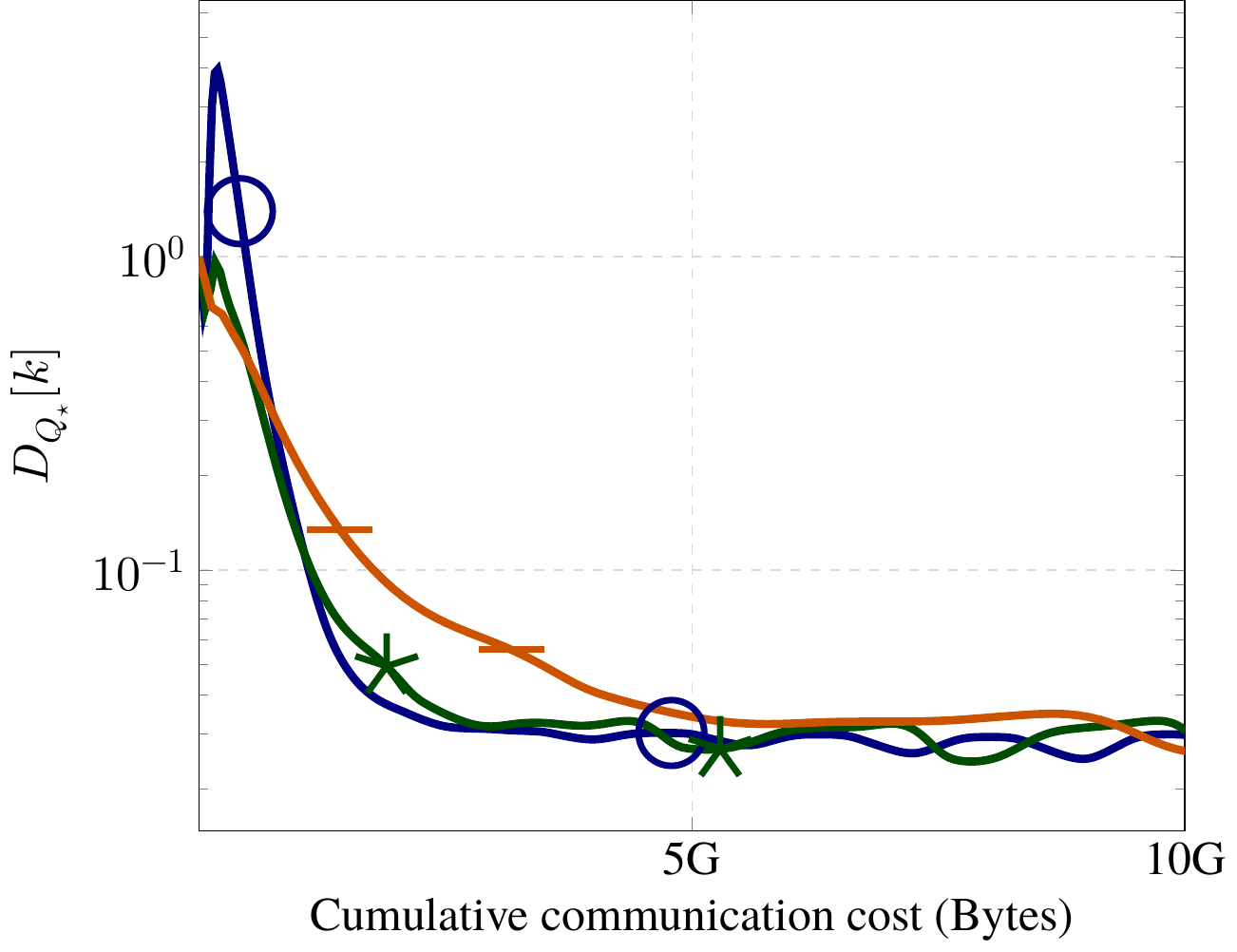}\label{fig:cartpole.distance.fixed.point}%
  }\\
  \subfloat[ Consensus losses~\eqref{consensus.loss} and \eqref{consensus.loss.nn} vs.\ $kM+m$, where the running
    indices are $(k, m)$: the outer VI index $k = 0, 1, \ldots$, and the inner consensus index $m = 0, \ldots, M-1$ ]{%
    \includegraphics[width = \RATIO\columnwidth]{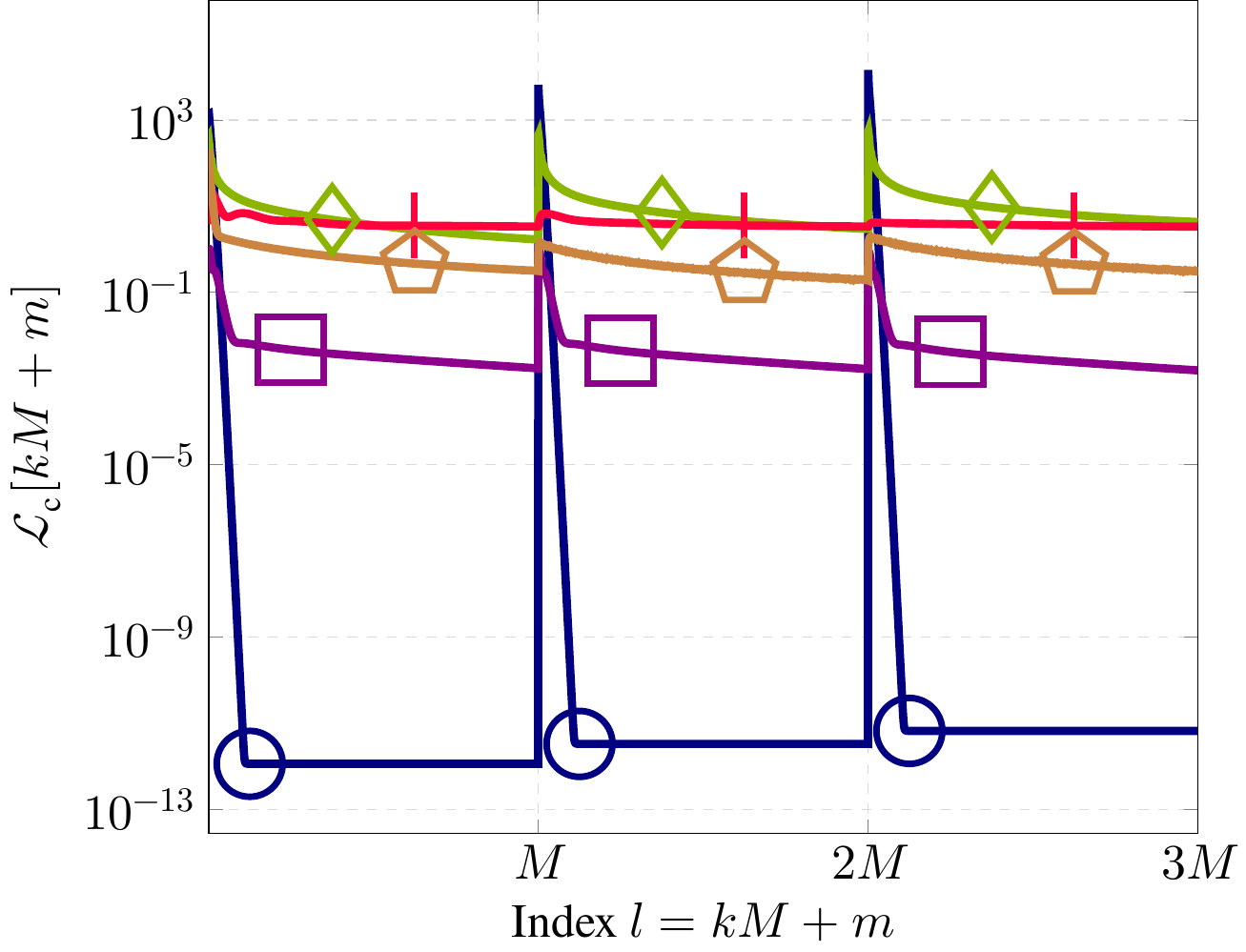}\label{fig:cartpole.consensus}%
  }
  \caption{Network of cartpoles (\cref{sec:cartpole}). D-FQ~\cite{distributed.TD}:~\protect \tikz[baseline = \BsLINE ex]
    { \protect \node[mark size = \MkSIZE pt, color = darkmagenta ] {\protect \pgfuseplotmark{square}}; }\,,
    D-LSTD~\cite{distributed.LSTD}:~\protect \tikz[baseline = \BsLINE ex]{ \protect \node[mark size = \MkSIZE pt, color
        = americanrose ] {\protect \pgfuseplotmark{|}}}, Gossip-NN~\cite{NN.gossip} (NN with \num{218}
    parameters):~\protect \tikz[baseline = \BsLINE ex]{ \protect \node[mark size = \MkSIZE pt, color = peru ] {\protect
        \pgfuseplotmark{pentagon}}}\,, Gossip-NN~\cite{NN.gossip} (NN with \num{386} parameters):~\protect
    \tikz[baseline = \BsLINE ex]{ \protect \node[mark size = \MkSIZE pt, color = black ] {\protect
        \pgfuseplotmark{x}}}\,, D-TD[ADMM]:~\protect \tikz[baseline = \BsLINE ex]{ \protect \node[mark size = \MkSIZE
        pt, color = applegreen ] {\protect \pgfuseplotmark{diamond}}}\,, \cref{algo} $(D, J_C) = (250, 50)$:~\protect
    \tikz[baseline = \BsLINE ex]{ \protect \node[mark size = \MkSIZE pt, color = navy ] {\protect
        \pgfuseplotmark{o}}}\,, \cref{algo} $(D, J_C) = (250, 25)$:~\protect \tikz[baseline = \BsLINE ex]{ \protect
      \node[mark size = \MkSIZE pt, color = darkgreen ] {\protect \pgfuseplotmark{star}}}\,, \cref{algo} $(D, J_C) =
    (250, 10)$:~\protect \tikz[baseline = \BsLINE ex]{ \protect \node[mark size = \MkSIZE pt, color = burntorange ]
      {\protect \pgfuseplotmark{-}}}\,, \cref{algo} $(D, J_C) = (150, 50)$:~\protect \tikz[baseline = \BsLINE ex]{
      \protect \node[mark size = \MkSIZE pt, color = maroon ] {\protect \pgfuseplotmark{triangle}}}\,. The shaded areas
    in \cref{fig:cartpole.episodic} correspond to values in the range of $(\text{mean}) \pm 0.5 \times (\text{standard
      deviation})$. The $M$-periodic jumps observed in \cref{fig:cartpole.consensus} occur because the algorithms
    perform $M$ consensus steps (index $m$) across the graph before each VI update (index
    $k$).} \label{fig:cartpole.losses}
\end{figure}

The data trajectory $\symcal{T}^{(n)}$ of \cref{ass:setting.trajectory}, with $N_\textnormal{av}^{(n)} = 100$, is
generated here in a similar way to \cref{sec:pendulum}. In other words, to mimic realistic scenarios when using the
transition module $F_{ \text{trans} } (\cdot)$ of \cite{openai, cartpole.software} to update $( x_{i+1}^{(n)},
v_{i+1}^{(n)}, \theta_{i+1}^{(n)}, \dot{\theta}_{i+1}^{(n)} ) \coloneqq F_{ \text{trans} } ( x_i^{(n)} + \epsilon_1,
v_i^{(n)} + \epsilon_2, \theta_i^{(n)} + \epsilon_3, \dot{\theta}_i^{(n)} + \epsilon_4, a_i^{(n)} + \epsilon_5 )$, noise
$\epsilon_k$ that follows the Gaussian PDF $\symcal{N}(0, \sigma_k^2)$, $k\in \{1, \ldots, 5\}$, is added, with
$\sigma_1^2 = 0.05$, $\sigma_2^2 = 0.5$, $\sigma_3^2 = 0.05$, $\sigma_4^2 = 0.5$, and $\sigma_5^2 = 0.05$. The only
twist here is that in the case where $\lvert x_{i+1}^{(n)} \rvert > B_x$ or $\lvert \theta_{i+1}^{(n)} \rvert >
B_{\theta}$, then $( x_{i+1}^{(n)}, v_{i+1}^{(n)}, \theta_{i+1}^{(n)}, \dot{\theta}_{i+1}^{(n)} )$ is redefined as the
initial $( x_0^{(n)}, v_0^{(n)}, \theta_0^{(n)}, \dot{\theta}_0^{(n)} )$ which is provided in~\cite{openai}.

In the current scenario, $N_{\text{e}} = 500$ in~\eqref{episodic.loss}, the RFF dimension $D = 250$ in~\eqref{RFF} for
all employed methods ($D = 150$ is also tested in \cref{fig:cartpole.episodic} for \cref{algo}), and $\sigma = 0.025$
in~\eqref{psi.n.k}. All other parameters of \cref{algo} are kept the same as in \cref{sec:pendulum}.

\cref{fig:cartpole.losses} reveals similar observations to those made at the end of
\cref{sec:pendulum}. However, D-LSTD seems to perform better in \cref{fig:cartpole.episodic}
compared to \cref{fig:pendulum.episodic}. Additionally, the differences in convergence speed between
the curves in \cref{fig:cartpole.distance.fixed.point} are less pronounced than those in
\cref{fig:pendulum.distance.fixed.point}.

\section{Conclusions}\label{sec:conclusions}

A novel class of nonparametric Bellman mappings (B-Maps) was introduced for value iteration (VI) in distributed
reinforcement learning (DRL). This approach leveraged a reproducing kernel Hilbert space representation of the
Q-function, enabling a nonparametric formulation that supports flexible, agent-specific basis function design. Beyond
sharing Q-functions, agents also exchanged basis information without relying on a centralized node, facilitating
consensus. The proposed methodology was backed by rigorous theoretical analysis, and numerical evaluations on two
well-known control problems demonstrated its superior performance compared to existing methods. Interestingly, the
evaluations revealed a counter-intuitive insight: despite involving increased information exchange---specifically
through covariance matrix sharing---the approach achieved the desired performance with lower cumulative communication
cost than prior-art DRL schemes. This underscores the critical role of basis information in accelerating the learning
process.

Ongoing research aims to extend this framework in several directions. In particular, future work will investigate the
extension of \cref{ass:setting} to MARL and multi-task RL, including scenarios that permit the sharing of state-action
information among agents; address the specific federated RL problem and its idiosyncrasies within a star-topology
network; consider online and streaming data scenarios; provide a theoretical analysis of the approximation error
introduced by the RFF approximation; and develop strategies to reduce the computational complexity of the matrix
inversion in~\eqref{psi.n.k}.

%%%%%%%%%%%%%%%%%%%%%%%%%%%%%%%%%%%%%%%%%%%%%%%%%%%%%%%%%%%%%%%%%

\appendix

The discussion starts with the following lemma to establish properties on recursions~\eqref{ACFB.Q}
and~\eqref{ACFB.C}. To save space, those recursions are unified in the generic form of~\eqref{FM-HSDM.x}.

\begin{lemma}\label{lem:conv.rate.general}
  For the user-defined
  $\vect{x}_{-1}, \vect{x}^{\prime} \in \Real^N$, generate
  sequence $( \vect{x}_m )_{m\in \IntegerP} \subset \Real^N$
  by%
  \begin{subequations}\label{FM-HSDM.x}%
    \begin{alignat}{2}
      \vect{x}_0
      & {} \coloneqq {}
      && A_{\varpi} ( \vect{x}_{-1}) - \eta(
         \vect{x}_{-1} - N\,
         \vect{x}^{\prime}) \label{FM-HSDM.initial.x}
      \\
      \vect{x}_{m+1}
      & \coloneqq
      && \vect{x}_{m} - (\, A_{\varpi}
         ( \vect{x}_{m-1} ) - \eta\,
         \vect{x}_{m-1}\, ) \notag \\
      &&& \phantom{ \vect{x}_{m} } + (\, A(\vect{x}_{m})
          - \eta\, \vect{x}_m\, )\,, \quad \forall
          m\in\IntegerP\,, \label{FM-HSDM.iteration.x}
    \end{alignat}
  \end{subequations}%
  where $\varpi\in [1/2, 1)$, $\eta\in (0, 2(1-\varpi)\, )$, $A \coloneqq \vect{I}_N - \gamma \vect{L}$, with $\vect{L}$
    being the $N\times N$ graph Laplacian matrix, and $A_{\varpi} \coloneqq \varpi A + (1-\varpi) \vect{I}_N$.  Then, $0
    < \varrho(\eta) < 1$ (\cref{thm:varrho}), and
  \begin{align}
    \norm{ \vect{x}_m - \vect{x}_* } = \symcal{O} (\, m\,
    \varrho^m (\eta)\, ) \,, \label{converge.rate.x}
  \end{align}
  where
  $\vect{x}_* \coloneqq [\vect{1}_N^{\intercal}
  \vect{x}^{\prime}, \ldots, \vect{1}_N^{\intercal}
  \vect{x}^{\prime} ]^{\intercal} = \vect{1}_{N\times N}\,
  \vect{x}^{\prime} \in \Real^N$.
\end{lemma}

\begin{IEEEproof}
  Notice that~\eqref{FM-HSDM.x} can be recast as%
  \begin{subequations}\label{FM-HSDM.x.form1}%
    \begin{align}%
      \vect{x}_0
      & = [ \varpi (\vect{I}_N - \gamma \vect{L} ) + (1 -
        \varpi) \vect{I}_N - \eta \vect{I}_N ] \vect{x}_{-1} +
        \eta N
        \vect{x}^{\prime}\,, \label{FM-HSDM.initial.x.form1}
    \end{align}
    and%
    \begin{alignat}{2}
      &&& \negphantom{ {} = {} } \vect{x}_{m+1} \notag\\
      & {} = {}
        && \vect{x}_{m} - (\, \varpi (\vect{I}_N
           - \gamma \vect{L} ) \vect{x}_{m-1} +
           (1-\varpi) \vect{x}_{m-1} -
           \eta\, \vect{x}_{m-1}\, ) \notag\\
      &&& \phantom{ \vect{x}_{m} } + (\, (\vect{I}_N -
          \gamma \vect{L} ) \vect{x}_{m}- \eta\, \vect{x}_{m}\,
          ) \notag \\
      & =
        && (\, (2-\eta) \vect{I}_N - \gamma \vect{L}\, )
           \vect{x}_m + (\, \varpi
           \gamma \vect{L} + (\eta-1) \vect{I}_N\, )
           \vect{x}_{m-1}\,. \label{FM-HSDM.iteration.x.form1}
    \end{alignat}%
  \end{subequations}%
  Define $\vect{a}_m \coloneqq \vect{U}^{\intercal} \vect{x}_m$ and $\vect{a}^{\prime} \coloneqq \vect{U}^{\intercal}
  \vect{x}^{\prime}$, where $\vect{U}$ is obtained by the EVD of $\vect{L}$, and let $a_m^{(n)}$ and $a^{\prime (n)}$ be
  the $n$th entries of $\vect{a}_m$ and $\vect{a}^{\prime}$, respectively. Applying $\vect{U}^{\intercal}$
  to~\eqref{FM-HSDM.x.form1} yields that $\forall n\in \symcal{N} \coloneqq \{1, \ldots, N\}$,%
  \begin{subequations}
    \begin{align}
      a_0^{(n)}
      & = -q_n a_{-1}^{(n)} + \eta N a^{\prime (n)}
        \,, \label{FM-HSDM.initial.a}
        \intertext{and $\forall m\in \IntegerP$,}
        a_{m+1}^{(n)}
      & = ( 2 -\eta -\gamma \lambda_n )
        a_m^{(n)} + ( \varpi\gamma \lambda_n + \eta -1 )
        a_{m-1}^{(n)} \notag \\
      & = p_n a_m^{(n)} + q_n a_{m-1}^{(n)}
        \,. \label{FM-HSDM.iteration.a}
    \end{align}
  \end{subequations}
  Let
  \begin{align*}
    \theta_n^{+}
    \coloneqq \frac{ p_n + \sqrt{p_n^2 + 4q_n} }{2}\,, \quad
    \theta_n^{-}
    \coloneqq \frac{ p_n - \sqrt{p_n^2 + 4q_n} }{2}\,,
  \end{align*}
  be the solutions of the quadratic equation $\theta^2 - p_n\theta - q_n = 0$, so that
  $p_n = \theta_n^{+} + \theta_n^{-}$ and $q_n = - \theta_n^{+} \theta_n^{-}$. As such,
  \eqref{FM-HSDM.iteration.a} yields%
  \begin{alignat}{2}
    \overbrace{ a_{m+1}^{(n)} - \theta_n^{+} a_{m}^{(n)} }^{ \Delta_{m+1}^{(n)+} }
    & {} = {}
    && \theta_n^{-} \overbrace{ ( a_{m}^{(n)} - \theta_n^{+}
       a_{m-1}^{(n)}) }^{ \Delta_m^{(n)+} }
       \,, \label{consecutive.Delta.plus} \\
    \underbrace{ a_{m+1}^{(n)} - \theta_n^{-} a_{m}^{(n)} }_{ \Delta_{m+1}^{(n)-} }
    & =
    && \theta_n^{+} \underbrace{ (a_{m}^{(n)} - \theta_n^{-}
       a_{m-1}^{(n)}) }_{ \Delta_m^{(n)-} } \,, \notag
  \end{alignat}
  which lead by induction to the following: $\forall m\in
  \IntegerPP$, $\forall n\in \symcal{N}$,%
  \begin{subequations}%
    \begin{alignat}{2}
      \Delta_{m}^{(n)+}
      & {} = {}
      && ( \theta_n^{-} )^{m}
         \Delta_0^{(n)+}\,, \label{FM-HSDM.iteration.Delta.a}
      \\
      \Delta_{m}^{(n)-}
      & =
      && ( \theta_n^{+} )^{m} \Delta_0^{(n)-}
         \,, \label{FM-HSDM.iteration.Delta.b}
         \intertext{with}
         \Delta_0^{(n)+}
      & = && - ( q_n + \theta_n^+ ) a_{-1}^{(n)} +
             \eta N a^{\prime (n)}\,, \label{Delta.0+}\\
      \Delta_0^{(n)-}
      & = && - ( q_n + \theta_n^- ) a_{-1}^{(n)} +
             \eta N a^{\prime (n)}\,. \label{Delta.0-}
    \end{alignat}%
  \end{subequations}

  The case of $n \in \symcal{N} \setminus \{N\}$ will be now considered. Recall that in this case $\lambda_n >
  0$. First, does there exist an $n\in \symcal{N} \setminus \{N\}$ s.t.\ $\theta_n^{+} = 1$? The answer is negative. To
  see this, assume for a contradiction that $\theta_n^{+} = 1$ for some $n\in \symcal{N} \setminus \{N\}$. Then, because
  $p_n = \theta_n^{+} + \theta_n^{-}$,
  \begin{alignat*}{2}
    && p_n & = 1 + \theta_n^{-} \\
    \Rightarrow {}
    && 2 - 2\eta - 2\gamma\lambda_n
          & = 2 \theta_n^{-} = p_n - \sqrt{p_n^2 + 4q_n}
    \\
    \Rightarrow {}
    && \eta + \gamma \lambda_n
          & = \sqrt{p_n^2 + 4q_n} \\
    \Rightarrow {}
    && ( \eta + \gamma \lambda_n )^2
      & = 4 + ( \eta + \gamma \lambda_n )^2 - 4 ( \eta +
         \gamma \lambda_n ) + 4q_n \\
    \Rightarrow {}
    && \gamma \lambda_n
          & = \varpi \gamma \lambda_n \quad ( \gamma
            \lambda_n \neq 0 )\\
    \Rightarrow {}
    && 1 & = \varpi \,,
  \end{alignat*}
  which contradicts the original design $\varpi < 1$.

  It has been already noted by the discussion after~\eqref{ACFB.Q} that sequence $(\vect{x}_m)_{m\in \IntegerP}$
  converges. Hence, $( \vect{a}_m = \vect{U}^{\intercal} \vect{x}_m )_{m\in \IntegerP}$ converges $\Rightarrow (
  \Delta_m^{(n)-} )_{ m\in \IntegerP }$ converges $\Rightarrow ( \Delta_m^{(n)-} )_{ m\in \IntegerP }$ is a Cauchy
  sequence $\Rightarrow \lvert \Delta_{m+1}^{(n)-} - \Delta_m^{(n)-} \rvert$ converges to zero. Because $\vect{x}_{-1}$
  can be arbitrarily fixed, it can be chosen so that $\Delta_0^{(n)-} \neq 0$, $\forall n\in \symcal{N} \setminus
  \{N\}$; see~\eqref{Delta.0-}. Notice now that $\lvert \Delta_{m+1}^{(n)-} - \Delta_m^{(n)-} \rvert = \lvert
  \theta_n^{+} \rvert^{m}\, \lvert \theta_n^{+} - 1 \rvert\, \lvert \Delta_0^{(n)-} \rvert$, which suggests that
  $\forall n\in \symcal{N} \setminus \{N\}$, $\lvert \theta_n^{+} \rvert^{m} = \lvert \Delta_{m+1}^{(n)-} -
  \Delta_m^{(n)-} \rvert / ( \lvert \theta_n^{+} - 1 \rvert\, \lvert \Delta_0^{(n)-} \rvert )$ converges to zero, and
  this is feasible only if $\lvert \theta_n^{+} \rvert < 1$. Observe also that $1 - \eta < 1$ to establish $0 < \varrho(
  \eta) < 1$ (\cref{thm:varrho}).

  Consider the case where $n \in \symcal{N} \setminus \{N\}$ and $p_n^2 + 4q_n \neq 0$. Then, $\theta_n^{-} \neq
  \theta_n^{+}$. Moreover, because $p_n = 2 - \eta - \gamma \lambda_n \geq 2 - 2(1 - \varpi) -1 \geq 2 - 1 - 1 = 0$, it
  can be verified that $\lvert \theta_n^{-} \rvert \leq \lvert \theta_n^{+}
  \rvert$. Multiplying~\eqref{FM-HSDM.iteration.Delta.a} by $\theta_n^{-}$ and \eqref{FM-HSDM.iteration.Delta.b} by
  $\theta_n^{+}$ and subtracting the resultant equations yield
  \begin{alignat*}{2}
    \lvert a_m^{(n)} \rvert
    & {} = {} && \frac{1}{ \lvert \theta_n^{+} -
                 \theta_n^{-} \rvert } \Bigl
                 \lvert (\theta_n^{+})^{m+1} \Delta_0^{(n)-} - (
                 \theta_n^{-} )^{m+1} \Delta_0^{(n)+} \Bigr
                 \rvert \\
    & = && \frac{1}{ \lvert \theta_n^{+} -
           \theta_n^{-} \rvert } \Bigl
           \lvert (\theta_n^{+})^{m+1} \Delta_0^{(n)-} - (
           \theta_n^{-} )^{m+1} \Delta_0^{(n)-} \\
    &&& \phantom{ \frac{1}{ \lvert \theta_n^{+} -
        \theta_n^{-} \rvert } \Bigl
        \lvert } + ( \theta_n^{-} )^{m+1} \Delta_0^{(n)-} - (
        \theta_n^{-} )^{m+1} \Delta_0^{(n)+} \Bigr
        \rvert \\
    & \leq && \frac{ \lvert (\theta_n^{+})^{m+1} - (
              \theta_n^{-} )^{m+1}   \rvert }{ \lvert
              \theta_n^{+} - \theta_n^{-} \rvert } \lvert
              \Delta_0^{(n)-} \rvert \\
    &&& + \lvert ( \theta_n^{-} ) \rvert^{m+1} \frac{ \lvert
        \Delta_0^{(n)-} - \Delta_0^{(n)+} \rvert}{ \lvert
        \theta_n^{+} - \theta_n^{-} \rvert } \\
    & = && \left\lvert \sum\nolimits_{k=0}^m (\theta_n^{+})^{m-k}
           (\theta_n^{-})^k \right\rvert \lvert
           \Delta_0^{(n)-} \rvert \\
    &&& + \lvert ( \theta_n^{-} ) \rvert^{m+1} \frac{ \lvert
        a_{-1}^{(n)} ( \theta_n^{+} - \theta_n^{-} ) \rvert
        }{ \lvert \theta_n^{+} - \theta_n^{-} \rvert } \\
    & \leq && (m+1)\, \lvert ( \theta_n^{+} ) \rvert^m \,
              \lvert \Delta_0^{(n)-} \rvert + \lvert (
              \theta_n^{-} ) \rvert^m\, \lvert a_{-1}^{(n)}
              \rvert \\
    & \leq && 2m\, \lvert ( \theta_n^{+} ) \rvert^m\, \lvert
              \Delta_0^{(n)-} \rvert + m\, \lvert ( \theta_n^{+}
              ) \rvert^m\, \lvert a_{-1}^{(n)} \rvert \\
    & \leq && C_n\, m\, \varrho^m ( \eta ) \leq C\, m\,
              \varrho^m ( \eta ) \,,
  \end{alignat*}
  for some $C_n \in \RealPP$ and $C \coloneqq \max_{ n \in \symcal{N} \setminus \{N\} } C_n$. It is worth stressing here
  that $C_n$ and $C$ depend on $\vect{a}_{-1}$, $\vect{a}^{\prime}$, and hence on $\vect{x}_{-1}$ and
  $\vect{x}^{\prime}$. This delicate point will be addressed at~\eqref{before.induction.Q} via \cref{ass:bounded.seq}.

  Consider now the case where $n \in \symcal{N} \setminus \{N\}$ with $p_n^2 + 4q_n = 0$. Then, $\theta_n^{+} =
  \theta_n^{-} = p_n / 2$, and induction on~\eqref{FM-HSDM.iteration.Delta.a}, together with~\eqref{FM-HSDM.initial.a},
  yield
  \begin{alignat*}{2}
    a_m^{(n)}
    & {} = {}
    && \left( \frac{p_n}{2} \right)^m a_0^{(n)} + m
       \left( \frac{p_n}{2} \right)^m \Delta_0^{(n) +} \\
    & = && \left( \frac{p_n}{2} \right)^m \left[
           \left( \frac{p_n}{2} \right)^2 a_{-1}^{(n)} +
           \eta N a^{\prime (n)} \right] \\
    &&& + m \left( \frac{p_n}{2} \right)^m \left[
        \left( \frac{p_n}{2} \right) \left(
        \frac{p_n}{2} - 1 \right) a_{-1}^{(n)} +
           \eta N a^{\prime (n)} \right] \,.
  \end{alignat*}
  Because $\lvert p_n / 2 \rvert \leq \varrho(\eta)$, the previous result suggests that there exists $C_n\in \RealPP$
  s.t.\ $\lvert a_m^{(n)} \rvert \leq C_n\, m\, \varrho^m (\eta)$.

  Consider now the case of $n = N$. Recall that $\lambda_N = 0$ and the $N$th column of $\vect{U}$ is $\vect{1}_N/
  \sqrt{N}$. Then, $p_N = 2 - \eta$, $q_N = \eta - 1$, $p_N^2 + 4q_N = \eta^2$, $\theta_N^{+} = 1$, and $\theta_N^{-} =
  1 - \eta$. Notice also that the $N$th entry of vector $\vect{a}^{\prime}$ is $a^{\prime (N)} = (1 / \sqrt{N} )
  \vect{1}_N^{\intercal} \vect{x}^{\prime}$. Now, adding~\eqref{FM-HSDM.initial.a} to copies
  of~\eqref{consecutive.Delta.plus} for consecutive values of $m$ yields
  \begin{align*}
    & a_m^{(N)} \\
    & = ( 1 - \eta ) a_{-1}^{(N)} + \eta N a^{\prime (N)} +
      ( 1 - \eta ) \sum_{k=0}^{m-1} \Delta_k^{(N) +} \\
    & = ( 1 - \eta ) a_{-1}^{(N)} + \eta N a^{\prime (N)} +
      ( 1 - \eta ) \sum_{k=0}^{m-1} ( 1 - \eta )^k
      \Delta_0^{(N) +} \\
    & = N a^{\prime (N)} + (1-\eta)^{m+1} (\, a_{-1}^{(N)} - N
      a^{\prime (N)}\, ) \\
    & = \sqrt{N}\, \vect{1}_N^{\intercal}
      \vect{x}^{\prime} + (1-\eta)^{m+1} (\, a_{-1}^{(N)} -
      \sqrt{N}\, \vect{1}_N^{\intercal} \vect{x}^{\prime}\, )
      \,.
  \end{align*}
  Therefore, there exists $C_N\in \RealPP$ s.t.\
  \begin{align*}
    \lvert  a_m^{(N)} - \sqrt{N}\, \vect{1}_N^{\intercal}
      \vect{x}^{\prime} \rvert & = (1-\eta)^{m+1}\, \lvert a_{-1}^{(N)} -
      \sqrt{N}\, \vect{1}_N^{\intercal} \vect{x}^{\prime}
      \rvert \\
    & \leq (1-\eta)^{m+1} ( \lvert a_{-1}^{(N)} \rvert + N
      \norm{ \vect{x}^{\prime} } )\\
    & \leq C_N\, (1-\eta)^{m+1} \leq C_N\, (1-\eta)^m \\
    & \leq C_N\, \varrho^m(\eta) \leq C_N\, m\,
      \varrho^m(\eta) \,.
  \end{align*}

  To summarize all of the previous findings, recall that $\vect{1}_N/ \sqrt{N}$ is the $N$th column of the orthogonal
  $\vect{U}$, so that $\vect{U}^{\intercal} \vect{1}_N = [ 0, 0, \ldots, \sqrt{N} ]^{\intercal}$, and
  \begin{alignat*}{2}
    \vect{a}_*
    & {} \coloneqq {}
    && \vect{U}^{\intercal} \vect{x}_* =
       \vect{U}^{\intercal} [ \vect{1}_N, \ldots, \vect{1}_N]
           \vect{x}^{\prime}\\
    & = && \left[ \begin{smallmatrix}
      \vect{0}^{\intercal} \\
      \vdots \\
      \vect{0}^{\intercal} \\
      \sqrt{N}\, \vect{1}_N^{\intercal}
    \end{smallmatrix} \right] \vect{x}^{\prime} =
           \left[ \begin{smallmatrix}
             0 \\
             \vdots \\
             0 \\
             \sqrt{N}\, \vect{1}_N^{\intercal}
             \vect{x}^{\prime}
           \end{smallmatrix} \right] \,.
  \end{alignat*}
  Therefore, there exists $C\in \RealPP$ s.t.\
  \begin{align*}
    \norm{ \vect{x}_m - \vect{x}_* }^2
    & = \norm{ \vect{U} ( \vect{x}_m - \vect{x}_*)
      }^2 = \norm{ \vect{a}_m - \vect{a}_* }^2 \\
    & = \sum\nolimits_{ n\in \mathcal{N} \setminus \{N\} }
      \lvert a_m^{(n)} \rvert^2 + \lvert a_m^{(N)} -
      \sqrt{N}\, \vect{1}_N^{\intercal} \vect{x}^{\prime}
      \rvert^2 \\
    & \leq C\, m^2 \varrho^{2m} (\eta) \,,
  \end{align*}
  which establishes~\eqref{converge.rate.x}.
\end{IEEEproof}

Now, by applying the transposition operator $\intercal$ to~\eqref{ACFB.Q} and by recalling that $\vect{L}$ is symmetric,
it can be verified that~\eqref{ACFB.Q} can be viewed as~\eqref{FM-HSDM.x}, where $\vect{x}_m$ refers to the $d$th column
$\vect{q}_m^{(d)}[k]$ ($d\in \{1, \ldots, D\}$) of the $N\times D$ matrix $\symbffrak{Q}_m^{\intercal}[k]$,
$\vect{x}_{-1}$ refers to the $d$th column of $\symbffrak{Q}_{-1}^{\intercal}[k]$, $\vect{x}^{\prime}$ to the $d$th
column $\vect{q}_{T}^{(d)}[k]$ of $\vect{T}^{\intercal} ( \symbffrak{Q} [k] )$, $\vect{x}_* = \vect{1}_{N\times N}\,
\vect{q}_{T}^{(d)}[k]$, and $A \coloneqq A^{ \textnormal{Q} }$. Then, by stacking together all of the aforementioned $D$
columns into matrices and by applying \cref{lem:conv.rate.general}, it can be verified that there exists $C\in \RealPP$
s.t.\
\begin{align*}
  C\, m^2 \varrho^{2m} (\eta)
  & \geq \sum_{ d\in \{1, \ldots, D\} } \norm{\,
    \vect{q}_m^{(d)}[k] - \vect{1}_{N\times N}\,
    \vect{q}_{T}^{(d)}[k]\, }^2 \\
  & = \norm{\, \symbffrak{Q}_m^{\intercal}[k] -
    \vect{1}_{N\times N}\, [\, \vect{q}_{T}^{(1)} [k],
    \ldots, \vect{q}_{T}^{(D)}[k]\, ]\,
    }_{\textnormal{F}}^2 \\
  & = \norm{\, \symbffrak{Q}_m^{\intercal}[k] -
    \vect{1}_{N\times N}\, \vect{T}^{\intercal}(
    \symbffrak{Q}[k] )\, }_{\textnormal{F}}^2  \\
  & = \norm{\,
    \symbffrak{Q}_m[k] - \vect{T}( \symbffrak{Q}[k] )\,
    \vect{1}_{N\times N}\, }_{\textnormal{F}}^2 \\
  & = \sum\nolimits_{ n\in \symcal{N} } \norm{\, Q_m^{(n)}[k] -
    \vect{T}( \symbffrak{Q} [k] )\, \vect{1}_N\, }^2 \\
  & \geq \norm{\, Q_m^{(n)}[k] - \vect{T}( \symbffrak{Q} [k]
    )\, \vect{1}_N\, }^2 \,,
\end{align*}
which establishes \cref{thm:Q.consensus.rate}. A similar sequence of arguments leads to the proof of
\cref{thm:C.consensus.rate}.

To prove \cref{thm:conv.Q.star}, recall first that $Q^{(n)} [k+1] = Q_M^{(n)} [k]$ from line~\ref{algo:VI.update} of
\cref{algo}, and that $Q_{\odot} = T_{\odot}( Q_{\odot} )$ by definition. Then,
\begin{alignat}{2}
  &&& \negphantom{ {} = {} } \norm{\, Q^{(n)} [k+1] -
      Q_{\odot}\, } \notag \\
  & {} = {} && \norm{\, Q_M^{(n)}[k] - Q_{\odot}\, } \notag \\
  & \leq && \norm{\, T_{\odot} ( Q^{(n)} [k]) - Q_{\odot}\, } + \norm{\, \vect{T}( \symbffrak{Q}[k] )\,
            \vect{1}_N - T_{\odot} ( Q^{(n)} [k])\, } \notag \\
  &&& + \norm{\, Q_M^{(n)} [k] - \vect{T} ( \symbffrak{Q}[k] )\, \vect{1}_N\, } \notag \\
  & \leq && \beta_{\odot} \, \norm{\,  Q^{(n)} [k] - Q_{\odot}\, } + \norm{\, \vect{T} ( \symbffrak{Q} [k] )\,
    \vect{1}_N - T_{\odot} ( Q^{(n)} [k])\, } \notag \\
  &&& + \norm{\, Q_M^{(n)} [k] - \vect{T} ( \symbffrak{Q} [k] )\, \vect{1}_N\, }\,, \label{k.recursion.i}
\end{alignat}
where the second inequality holds because of \cref{ass:contraction}. Observe now that
\begin{alignat*}{2}
  &&& \negphantom{ {} = {} } T_{\odot}( Q^{(n)}[k] ) \\
  & {} = {} && \sum\nolimits_{ n^{\prime} \in\symcal{N} } \vect{\Psi}_{\odot}^{(n^{\prime})}\,
               \vect{c}^{(n^{\prime})}( Q^{(n)}[k] ) \\
  & = && \sum\nolimits_{ n^{\prime} \in\symcal{N} } \vect{\Psi}^{(n^{\prime})}[k]\,
         \vect{c}^{(n^{\prime})}( Q^{(n)}[k] ) \\
  &&& + \sum\nolimits_{ n^{\prime} \in\symcal{N} } ( \vect{\Psi}_{\odot}^{(n^{\prime})} -
      \vect{\Psi}^{(n^{\prime})}[k] )\, \vect{c}^{(n^{\prime})}( Q^{(n)}[k] ) \\
  & = && \sum\nolimits_{ n^{\prime} \in\symcal{N} } T^{(n^{\prime})} ( Q^{(n)}[k] ) \\
  &&& + \sum\nolimits_{ n^{\prime} \in\symcal{N} } ( \vect{\Psi}_{\odot}^{(n^{\prime})} -
      \vect{\Psi}^{(n^{\prime})}[k] )\, \vect{c}^{(n^{\prime})}( Q^{(n)}[k] ) \,.
\end{alignat*}
An inspection of \eqref{psi.star.n} and \eqref{psi.n.k}, under the light of \cref{thm:C.consensus.rate}, and the
continuity of the mapping $( \cdot + \sigma \vect{I}_D )^{-1}$ suggest that for an arbitrarily fixed $\epsilon\in
\RealPP$ and for all sufficiently large $k$, $\norm{\, \vect{\Psi}_{\odot}^{(n^{\prime})} -
  \vect{\Psi}^{(n^{\prime})}[k]\, }_{ \textnormal{F} } \leq \epsilon$. Further, by \cref{ass:bounded.seq}, there exists
a $C^{\dprime}\in \RealPP$ such that for all sufficiently large $k$,
\begin{align*}
  \sum\nolimits_{ n^{\prime} \in\symcal{N} } \norm{\, ( \vect{\Psi}_{\odot}^{(n^{\prime})} -
      \vect{\Psi}^{(n^{\prime})}[k] )\, \vect{c}^{(n^{\prime})}( Q^{(n)}[k] )\, } \leq C^{\dprime}
  \epsilon\,.
\end{align*}
Via the previous observations,
\begin{alignat*}{2}
  &&& \negphantom{ {} = {} } \norm{\, \vect{T} ( \symbffrak{Q}
      [k] )\, \vect{1}_N - T_{\odot} (Q^{(n)}[k])\, } \\
  & {} \leq {} && \norm{\, \sum\nolimits_{n^{ \prime} \in \symcal{N} }
    T^{(n^{\prime})} (Q^{(n^{\prime})}[k]) -
    \sum\nolimits_{n^{\prime} \in \symcal{N} }
                  T^{(n^{\prime})} (Q^{(n)}[k])\, } \\
  &&& + \sum\nolimits_{ n^{\prime} \in\symcal{N} } \norm{\, ( \vect{\Psi}_{\odot}^{(n^{\prime})} -
      \vect{\Psi}^{(n^{\prime})}[k] )\, \vect{c}^{(n^{\prime})}( Q^{(n)}[k] )\, } \\
  & \leq && \sum\nolimits_{ n^{\prime} \in \symcal{N} }
    \norm{\, T^{(n^{\prime})} ( Q^{(n^{\prime})}[k] ) -
    T^{(n^{\prime})} (Q^{(n)}[k])\, } + C^{\dprime} \epsilon \\
  & \leq && \sum\nolimits_{n^{\prime} \in \symcal{N}}
    \beta^{(n^{\prime})}[k]\, \norm{\, Q^{(n^{\prime})}[k] -
    Q^{(n)}[k]\, } + C^{\dprime} \epsilon \\
  & \leq && \sum_{n^{\prime} \in\symcal{N}}
    \beta^{(n^{\prime})}[k]\, \norm{\, Q^{(n^{\prime})}[k] -
    \vect{T} ( \symbffrak{Q} [k-1] )\, \vect{1}_N \, } \\
  &&& + \sum_{n^{\prime} \in\symcal{N}}
    \beta^{(n^{\prime})}[k]\, \norm{\, \vect{T} ( \symbffrak{Q} [k-1]
      )\, \vect{1}_N - Q^{(n)}[k]\, } + C^{\dprime} \epsilon \\
  & \leq && C^{\prime} \sum\nolimits_{n^{\prime}
            \in\symcal{N}} \norm{\, Q_M^{(n^{\prime})} [k-1] -
            \vect{T} ( \symbffrak{Q} [k-1] )\, \vect{1}_N \, } \\
  &&& + N C^{\prime} \norm{\, \vect{T} ( \symbffrak{Q} [k-1] )\,
      \vect{1}_N - Q_M^{(n)}[k-1]\, } + C^{\dprime} \epsilon \,,
\end{alignat*}
where the existence of $C^{\prime}$ is guaranteed by \cref{ass:bounded.Lipschitz}. Therefore, \eqref{k.recursion.i}
becomes
\begin{alignat}{2}
  &&& \negphantom{ {} \leq {} } \norm{\, Q^{(n)} [k+1] -
      Q_{\odot}\, } \notag \\
  & {} \leq {}
    && \beta_{\odot}\, \norm{\,  Q^{(n)} [k] - Q_{\odot}\, }
  \notag \\
  &&& + C^{\prime} \sum\nolimits_{n^{\prime}
            \in\symcal{N}} \norm{\, Q_M^{(n^{\prime})} [k-1] -
            \vect{T}( \symbffrak{Q}[k-1] )\, \vect{1}_N \, }
      \notag \\
  &&& + N C^{\prime} \norm{\, \vect{T} ( \symbffrak{Q} [k-1] )\,
      \vect{1}_N - Q_M^{(n)}[k-1]\, } + C^{\dprime} \epsilon \notag \\
  &&& + \norm{\, Q_M^{(n)} [k] - \vect{T} ( \symbffrak{Q} [k] )\,
      \vect{1}_N\, } \notag \\
  & \leq
    && \beta_{\odot}\, \norm{\,  Q^{(n)} [k] - Q_{\odot}\, }
       + C\, ( M\, \varrho^M (\eta) + \epsilon) \,, \label{before.induction.Q}
\end{alignat}
for some $C\in \RealPP$, where the existence of $C$ is ensured by $C^{\dprime}$, \cref{thm:Q.consensus.rate} and
\cref{ass:bounded.seq}. Now, by using induction on~\eqref{before.induction.Q}, it can be verified that there exists a
sufficiently large $k_0\in \IntegerPP$ such that for all $\IntegerPP\ni k > k_0$,
\begin{alignat*}{2}
  \norm{\, Q^{(n)} [k + k_0] - Q_{\odot} \, }
  & {} \leq {}
    && \beta_{\odot}^k \norm{\, Q^{(n)} [k_0] - Q_{\odot}
       \, } \\
  &&& + C\, ( M\, \varrho^M (\eta) + \epsilon ) \sum\nolimits_{i=0}^{k-1} \beta_{\odot}^i\,.
\end{alignat*}
The application of $\lim\sup_{k\to \infty}$ to both sides of the previous inequality and the fact that $\epsilon \in
\RealPP$ was arbitrarily fixed establish \cref{thm:conv.Q.star}.

Moving on to the proof of \cref{thm:optimal.eta}, notice that under \cref{ass:gamma,ass:varpi}, $p_n + (p_n^2 +
4q_n)^{1/2} = 2 - \eta - b_n + (b_n^2 + 2 (\eta - 1) b_n + \eta^2 )^{1/2}$ in \eqref{varrho.def}.

\begin{lemma}\label{lemma:b.N-1.max}
  Let $\eta \in (0,1)$ and $b_{N-1} \in ( 0, 1/2 )$. Define the continuous function
  $f_{\eta} \colon (0,1] \to \Real \colon b \mapsto f_{\eta} (b) \coloneqq \lvert d_{\eta}(b)
  \rvert^2$, where
  \begin{align*}
    d_{\eta}(b) \coloneqq 2 - \eta - b + \sqrt{ b^2 + 2
    (\eta-1) b + \eta^2 } \,.
  \end{align*}
  Then, $\forall \eta\in (0,1)$, $b_{N-1} \in \arg
  \max_{\{\, b_n\, \given\, n \in \symcal{N} \setminus
    \{N\}\, \} } f_{\eta} (b_{n})$.
\end{lemma}

\begin{IEEEproof}
Notice that $\forall b \in (0,1]$, $2 - \eta - b > 2 - 1 - 1 = 0$. Consider first the case $\eta > 1/2$. Then $\forall b
\in (0,1]$, $b^2 + 2(\eta-1) b + \eta^2 > b^2 - b + 1/4 = (b - 1/2)^2 \geq 0 \Rightarrow d_{\eta}(b) > 0 \Rightarrow
  f_{\eta}(b) = d_{\eta}^2(b) \Rightarrow$
\begin{align}
  f_{\eta}^{\prime} (b) = 2\, d_{\eta}(b) \left (-1 +
  \frac{ b + \eta - 1}{ \sqrt{b^2 + 2 (\eta-1)b + \eta^2}}
  \right) \,. \label{f.differential}
\end{align}
Now, by
\begin{align}
  b^2 + 2 (\eta-1) b + \eta^2 = (b + \eta -1)^2 + 2 \eta -
  1 \,,\label{inequality.square}
\end{align}
and $\eta > 1/2$, it can be verified that $f_{\eta}^{\prime} (b) < 0$ in~\eqref{f.differential}. Hence $f_{\eta}(\cdot)$
is monotonically decreasing on $(0,1]$, and for any $n \in \symcal{N} \setminus \{N\}$, $f_{\eta}(b_{N-1}) \geq f_{\eta}
  (b_{n})$ because $b_{N-1}\leq b_{N-2} \leq \ldots \leq b_1 = 1$.

The following refer to the case $\eta \leq 1/2$. Define $x_1 \coloneqq 1 - \eta - \sqrt{1 - 2 \eta} > 0$ and $x_2
\coloneqq 1 - \eta + \sqrt{1 - 2 \eta}$, and notice that $x_1 < 1$ and $x_2 < 2(1 - \eta) < 2 -
\eta$. Use~\eqref{inequality.square} to verify that if $b \in (0, x_1) \Rightarrow b^2 + 2(\eta-1)b +\eta^2 > 0
\Rightarrow d_{\eta}(b) > 2 - \eta - b \geq 2 - 1/2 - b = 3/2 -b > 1/2 > 0 \Rightarrow f_{\eta}^{\prime}(b)$ is given
by~\eqref{f.differential}. Moreover, $b + \eta - 1 < x_1 + \eta -1 \leq 0 \Rightarrow f_{\eta}^{\prime}(b) < 0$. Hence,
$f_{\eta}(\cdot)$ is monotonically decreasing on $(0, x_1)$, and $f_{\eta} (b_{N-1}) \geq \max \{ f_{\eta} (b_n) \given
b_n\in (0, x_1), n \in \symcal{N} \setminus \{N\} \}$ whenever the latter set is nonempty.

If $b \in [ x_1, x_2] \cap (0,1]$, then by \eqref{inequality.square}, $b^2 + 2(\eta - 1)b + \eta^2 \leq 0 \Rightarrow
  f_{\eta} (b) = (2 - \eta -b)^2 - (b^2 + 2(\eta-1)b + \eta^2) = -2b + 4 - 4\eta$. Thus $f_{\eta} (\cdot)$ is
  monotonically decreasing on $[ x_1, x_2]$, and $f_{\eta} (b_{N-1}) \geq \max \{ f_{\eta} (b_n) \given b_n\in [ x_1,
    x_2], n \in \symcal{N} \setminus \{N\} \}$. Notice also that $b_{N-1} < 1/2 \leq 1 - \eta \leq x_2 \Rightarrow
  f_{\eta}(b_{N-1}) \geq f_{\eta}(1/2)$. These results hold true even if $x_2 \geq 1$.

Consider finally the case $x_2 < 1$. Extend function $f_{\eta}$ to the continuous $\bar{f}_{\eta} \colon (x_2, +\infty)
\to \Real \colon b \mapsto \bar{f}_{\eta} (b) \coloneqq \lvert d_{\eta}(b) \rvert^2$, so that $\bar{f}_{\eta}
\rvert_{(x_2, 1]} = f_{\eta}$. Notice by $b > x_2$, \eqref{inequality.square} and $2\eta - 1 \leq 0$ that $b +
  \eta - 1 > x_2 + \eta - 1 = (1 - 2\eta)^{1/2} \geq 0 \Rightarrow b^2 + 2 (\eta-1) b + \eta^2 > (x_2 + \eta - 1)^2 +
  2\eta - 1 = 1 - 2\eta + 2\eta - 1 = 0 \Rightarrow ( b + \eta - 1 ) / ( b^2 + 2 (\eta-1) b + \eta^2 )^{1/2} \geq 1$ and
  $\bar{f}_{\eta} (b) = d_{\eta}^2 (b)$.

The monotonicity of $\bar{f}_{\eta}(b)$ on $(x_2, +\infty)$ is going to be explored next. The case where $b\in ( x_2, 2
- \eta ]$ is considered first: $b\leq 2 - \eta \Rightarrow 2 - \eta - b \geq 0 \Rightarrow d_{\eta} (b) > 0 \Rightarrow
  \bar{f}_{\eta}^{\prime} (b) \geq 0$. In the case where $b > 2 - \eta$, notice from $4 - 4\eta - 2b < 4 - 4\eta - 2(2 -
  \eta) = -2\eta < 0$ and $2 - \eta - b < 0$ that
\begin{align*}
  d_{\eta} (b)
  & = \frac{ (2 - \eta - b)^2 - (b^2 + 2 (\eta-1) b + \eta^2) }{ 2 - \eta - b - \sqrt{
  b^2 + 2 (\eta-1) b + \eta^2 } } \\
  & = \frac{ 4 - 4\eta - 2b}{ 2 - \eta - b - \sqrt{ b^2 + 2 (\eta-1) b + \eta^2 } } > 0 \,.
\end{align*}
Hence, $\bar{f}_{\eta}^{\prime} (b) \geq 0$. To summarize, $\bar{f}_{\eta} (\cdot)$ is monotonically non-decreasing on
$(x_2, +\infty)$. Thus, $\forall b\in (x_2, +\infty)$, $\bar{f}_{\eta} (b) \leq \lim_{ b^{\prime} \to \infty }
\bar{f}_{\eta} (b^{\prime}) = \lim_{ b^{\prime} \to \infty } d_{\eta}^2 (b^{\prime}) = 1$. It has been already noted
earlier that $f_{\eta} (b_{N-1}) \geq f_{\eta} (1/2)$. Consequently, $\forall b\in (x_2, 1]$ and for $\eta \leq 1/2$,
  $f_{\eta} (b_{N-1}) \geq f_{\eta} (1/2) = d_{\eta}^2 (1/2) \geq (3/2 - \eta)^2 \geq 1 \geq \bar{f}_{\eta} (b) =
  f_{\eta} (b)$. This establishes $f_{\eta} (b_{N-1}) \geq \max \{ f_{\eta} (b_n) \given b_n\in (x_2, 1], n \in
    \symcal{N} \setminus \{N\} \}$ whenever the latter set is nonempty.
\end{IEEEproof}

\begin{lemma}\label{lemma:h.function}
  For $b \in (0, 1/2 )$, define $h_b \colon (0,1) \to \Real\colon \eta \mapsto h_b (\eta)$ as
  \begin{align*}
    h_b (\eta) \coloneqq
    \frac{ \left \lvert 2 - \eta - b + \sqrt{ \eta^2 +
    2(\eta-1)b + b^2 } \right \rvert^2 }{4}  \,.
  \end{align*}
  Then, $-b + \sqrt{2b} = \arg \min_{ \eta\in (0,1) } h_b
  (\eta)$.
\end{lemma}

\begin{IEEEproof}
Notice that $\eta \in (0,-b + (2b)^{1/2}\, ] \Rightarrow \eta^2 + 2(\eta-1)b + b^2 \leq 0 \Rightarrow h_b(\eta) = (1/4)
  (\, (2-\eta-b)^2 - (\eta^2+2b\eta+b^2-2b)\, ) = (1/2) (-2\eta +2 -b) \Rightarrow h_b^\prime (\eta) = -1 < 0$, $\forall
  \eta \in (0,-b + (2b)^{1/2}\, ]$.

Next, $\eta\in (-b + (2b)^{1/2}, 1) \Rightarrow \eta^2 + 2b \eta +b^2 -2b > 0$, and
\begin{align*}
  h_b^{\prime} (\eta) {} = {}
  & \frac{2 - \eta -b +\sqrt{\eta^2 + 2b\eta +b^2 -2b}}{2}
  \\
  & \cdot \Bigl ( -1 + \frac{\eta + b}{\sqrt{\eta^2 +
    2b\eta +b^2 - 2b}} \Bigr ) \,.
\end{align*}
Because $\eta + b >0$ and $\eta^2 + 2b \eta + b^2 -2b < (\eta + b)^2$, $(\eta+b) / ( \eta^2 + 2b \eta + b^2 - 2b )^{1/2}
> 1$. Moreover, $2 - \eta - b + (\eta^2 + 2b \eta + b^2 - 2b)^{1/2} > 0 \Rightarrow h_b^{\prime} (\eta) > 0$, $\forall
\eta \in (-b + (2b)^{1/2}, 1)$. Therefore, the claim of \cref{lemma:h.function} holds true.
\end{IEEEproof}

By \cref{lemma:h.function}, define $\eta_* \coloneqq - b_{N-1} + (2b_{N-1})^{1/2}$, and notice that $\forall \eta \in
(0,1)$,
\begin{align}
  \lvert 1 - ( 2b_{N-1} )^{1/2} / 2 \rvert = h_{b_{N-1}}^{1/2} ( \eta_* ) \leq h_{b_{N-1}}^{1/2} (\eta)
  \,. \label{last.ineq}
\end{align}
Moreover, by $b_{N-1} \in ( 0, 1/2 )$, $(2b_{N-1})^{1/2} / 2 \leq - b_{N-1} + (2b_{N-1})^{1/2} \leq 1$, and
\begin{align}
  h_{b_{N-1}}^{1/2} (\eta_*)
  & = \lvert 1 - (2b_{N-1})^{1/2} /2 \rvert \notag \\
  & \geq \lvert 1 - ( -b_{N-1} + (2b_{N-1})^{1/2} ) \rvert \notag \\
  & = \lvert 1 - \eta_* \rvert = 1 - \eta_*
    \,. \label{h.eta.star}
\end{align}

Observe now by the definitions of $f_{\eta}, h_b$ in \cref{lemma:b.N-1.max,lemma:h.function} that $\forall \eta\in
(0,1)$, $\forall b\in (0,1)$, $h_b(\eta) = f_{\eta}(b) / 4$ and
\begin{align}
  & \arg \max_{ b_n\, \given\, n\in \symcal{N} \setminus \{N\} } h_{b_n}^{1/2} (\eta) \notag \\
  & = \arg \max_{ b_n\, \given\, n\in \symcal{N} \setminus \{N\} } f_{\eta}^{1/2} (b_n)\, \ni b_{N-1}
  \,. \label{h.f.argmax}
\end{align}
Putting all arguments together, $\forall \eta \in (0,1)$,
\begin{alignat*}{2}
  \varrho (\eta_*)
  & = && \max \left \{ \max \{\, h_{b_n}^{1/2} (\eta_*) \given n\in
    \symcal{N} \setminus \{N\}\, \}, (1-\eta_*)
    \right \} \\
  & \overset{\eqref{h.f.argmax}}{=} && \max \left \{
    h_{b_{N-1}}^{1/2} (\eta_*), (1-\eta_*) \right \}
    \overset{\eqref{h.eta.star}}{=} h_{b_{N-1}}^{1/2}
    (\eta_*) \\
  & \leq && h_{b_{N-1}}^{1/2} (\eta) \leq \max \left \{
    h_{b_{N-1}}^{1/2} ( \eta ), (1 - \eta) \right \} \\
  & \overset{\eqref{h.f.argmax}}{=} && \max \left \{ \max \{\,
    h_{b_n}^{1/2} (\eta) \given n\in \symcal{N} \setminus
    \{N\}\, \}, (1-\eta) \right \} \\
  & = && \varrho(\eta) \,,
\end{alignat*}
where the first inequality holds because of \eqref{last.ineq}. The previous result establishes \cref{thm:optimal.eta}.

\printbibliography
\end{document}